\documentclass[10pt, a4paper]{article}

\usepackage{makecell}
\usepackage{multirow}
\usepackage{arydshln}
\usepackage{subcaption}

\usepackage[]{lrec2026}
\usepackage{cleveref}
\usepackage{graphicx}

\title{Piecing Together Cross-Document Coreference Resolution Datasets: Systematic Dataset Analysis and Unification}

\name{Anastasia Zhukova, Terry Ruas, Jan Philip Wahle, Bela Gipp} 

\address{University of G\"ottingen \\
         G\"ottingen, Germany\\
         {anastasia.zhukova, ruas, wahle, gipp}@uni-goettingen.de\\}

\abstract{
Work in Natural Language Understanding increasingly relies on the ability to identify and track entities and events across large, heterogeneous text collections. This task, known as cross-document coreference resolution (CDCR), has a wide range of downstream applications, including multi-document summarization, information retrieval, and knowledge base population.
Research in this area remains fragmented due to heterogeneous dataset formats, varying annotation standards, and the predominance of the CDCR definition as the event coreference resolution (ECR). To address these challenges, we introduce \textbf{uCDCR}, a unified dataset that consolidates diverse publicly available English CDCR corpora across various domains into a consistent format, which we analyze with standardized metrics and evaluation protocols. uCDCR incorporates both entity and event coreference, corrects known inconsistencies, and enriches datasets with missing attributes to facilitate reproducible research. We establish a cohesive framework for fair, interpretable, and cross-dataset analysis in CDCR and compare the datasets on their lexical properties, e.g., lexical composition of the annotated mentions, lexical diversity and ambiguity metrics, discuss the annotation rules and principles that lead to high lexical diversity, and examine how these metrics influence performance on the same-head-lemma baseline. Our dataset analysis shows that ECB+, the state-of-the-art benchmark for CDCR, has one of the lowest lexical diversities, and its CDCR complexity, measured by the same-head-lemma baseline, lies in the middle among all uCDCR datasets. Moreover, comparing document and mention distributions between ECB+ and uCDCR shows that using all uCDCR datasets for model training and evaluation will improve the generalizability of CDCR models. Finally, the almost identical performance on the same-head-lemma baseline, separately applied to events and entities, shows that resolving both types is a complex task and should not be steered toward ECR alone. The uCDCR dataset is available at \url{https://huggingface.co/datasets/AnZhu/uCDCR}, and the code for parsing, analyzing, and scoring the dataset is available at \url{https://github.com/anastasia-zhukova/uCDCR}.
 \\ \newline  
 }

\begin{document}

\maketitleabstract

\section{Introduction}
Cross-document coreference resolution (CDCR) is the task of identifying and linking expressions that refer to the same entities or events across multiple documents \cite{singh-etal-2011-large}, and remains a fundamental yet underexplored problem in natural language processing (NLP). For example, in the sentences "The President \textit{\underline{announced a new economic policy}} aimed at boosting the national economy" and "\textit{\underline{This initiative}} is expected to create thousands of new jobs across the country," the highlighted mentions refer to the same event. Despite recent progress in within-document coreference resolution and event extraction, CDCR still lacks standardized benchmarks and unified evaluation practices. 

\begin{figure}
    \centering
    \includegraphics[width=1.0\linewidth]{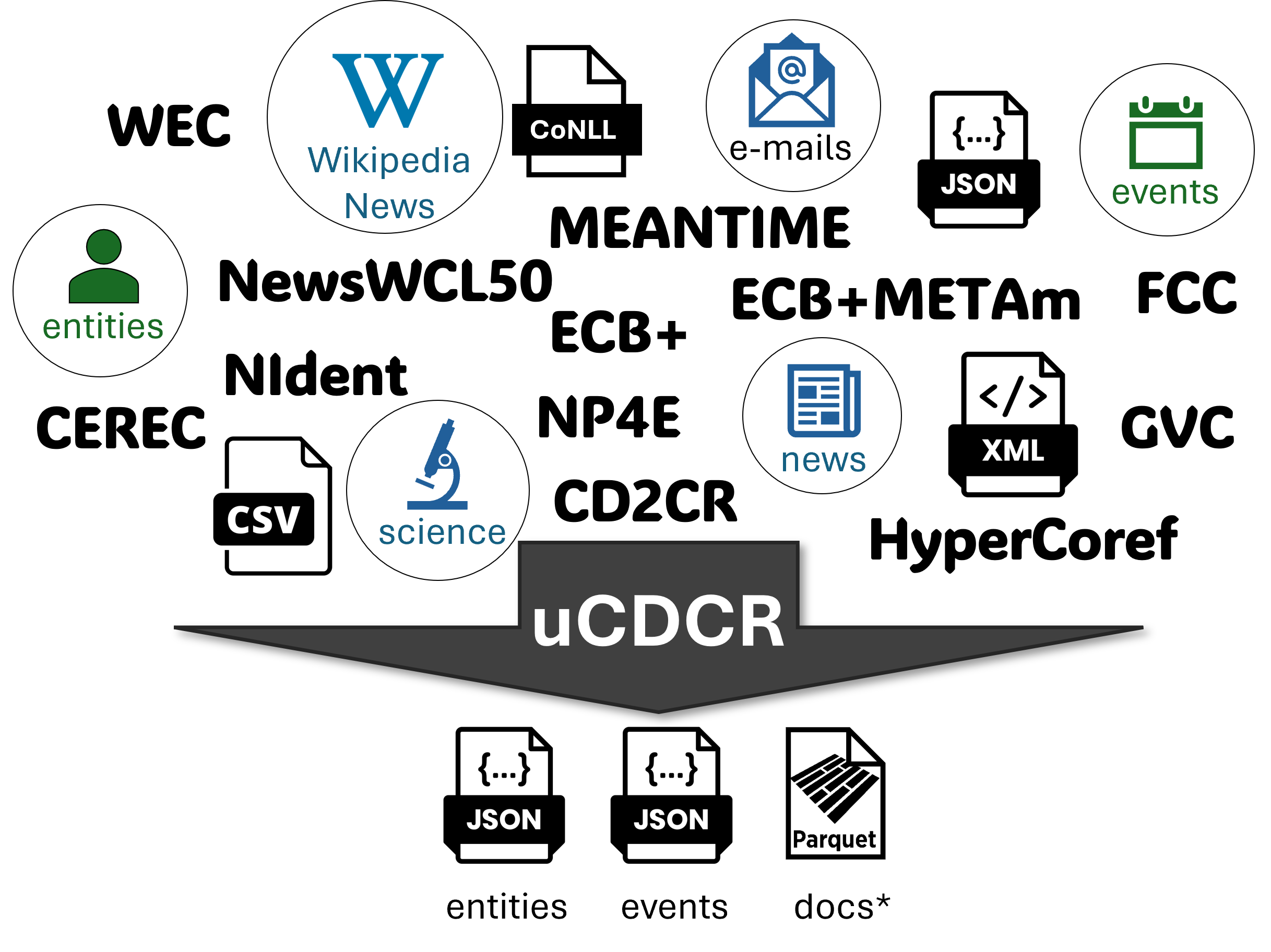}
    \caption{The existing CDCR datasets stem from the diverse annotation schemes, topics, and formats in which they were released. The proposed uCDCR dataset systemizes, unifies, and analyzes publicly available datasets using a single evaluation framework.}
    \label{fig:intro}
\end{figure}

Existing datasets vary widely in their annotation focus, including domains, topics, rules for mention detection, definitions of coreference links, and dataset formats \cite{zhukova-etal-2022-towards}. Moreover, CDCR has often been defined as cross-document event coreference resolution (ECR), as it is considered a more complex task than entity resolution \cite{hovy-etal-2013-events}, thereby downplaying the complexity of resolving entity mentions. As a result, most CDCR models are event-centric, and their performance is limited to either the ECB+ dataset \citep{cybulska-vossen-2014-using}, which serves as the state-of-the-art benchmark for CDCR, or to ECR-based datasets, thereby resulting in fragmented research progress in the field. This fragmentation contrasts sharply with other areas of NLP, where unified evaluation frameworks such as GLUE \citeplanguageresource{Wang2019} have catalyzed rapid model development and scientific reproducibility.

To address this gap, we introduce \textit{uCDCR, a unified CDCR dataset} that standardizes and consolidates evaluation across heterogeneous English datasets. uCDCR harmonizes diverse corpora spanning multiple domains, annotation schemes, topics, and dataset formats by reformatting them into a single JSON format from the XML-based, CoNLL, CSV, and JSON formats, and extracting the missing mention attributes, such as the head of the phrase and mention context, while preserving their unique annotation and structural characteristics (\Cref{fig:intro}). By establishing a common data format, metric suite, and evaluation protocol, uCDCR provides the first cohesive framework for fair and interpretable cross-dataset comparison in CDCR. Prior to this effort, the absence of such a unified benchmark hindered reproducibility and slowed methodological innovation.  Specifically, the main contributions of this paper are as follows:

\textbf{(1) Unified and comprehensive dataset collection}. We present uCDCR, the most extensive compilation of publicly available CDCR datasets to date, unifying both entity and event coreference resolution within a single framework. The dataset is released on the HuggingFace and GitHub to encourage broader participation in CDCR research beyond ECB+ and event-focused corpora.

\textbf{(2) Standardization and data correction}. All datasets have been standardized into a consistent, unified format through a harmonized parsing pipeline. This process includes re-parsing and correcting prior tokenization inconsistencies, ensuring reproducible token–text–token conversions for downstream analyses and correct mapping of mentions to their contexts or full documents. Additionally, missing attributes—such as topic and subtopic identifiers, as well as mention- and chain-level features—have been extracted and completed to provide a coherent structure across datasets.

\textbf{(3) Comprehensive dataset analysis framework.} We conduct a systematic analysis of all datasets in uCDCR using established metrics for lexical diversity and ambiguity, including the Measure of Textual Lexical Diversity (MTLD) \cite{McCarthy2010}, Phrasing Diversity (PD) \cite{zhukova-etal-2022-towards}, and lexical ambiguity \cite{eirew-etal-2021-wec} measures. We further evaluate dataset difficulty by comparing performance against the same-head-lemma baseline, providing insights into the relationships between lexical properties and CDCR model performance, and setting a baseline for model benchmarking.

\begin{table*}[]
\scriptsize
\centering
\begin{tabular}{|l|c|c|c|c|c|c|c|c|} 
\hline
\multirow{2}{*}{\textbf{Dataset}} & \multirow{2}{*}{\makecell[c]{\textbf{Mention}\\\textbf{type}}}  & \multirow{2}{*}{\textbf{Domain}}  & \multicolumn{3}{c|}{\makecell[c]{\textbf{Dataset collection and annotation}}} & \multirow{2}{*}{\makecell[c]{\textbf{Mention}\\ \textbf{span}}} & \multirow{2}{*}{\makecell[c]{\textbf{Annotated} \\\textbf{pronouns}}} & \multirow{2}{*}{\makecell[c]{\textbf{Near-}\\\textbf{identity}}} \\
\cline{4-6}
& & &subtopics & mentions & chains & & & \\
\hline
CD2CR & N & \makecell[l]{news \& science} & autom. & semi-aut. & semi-aut. & max & - & - \\
\hdashline
CEREC\textsubscript{exp} & N & emails & autom. & semi-aut. & semi-aut. & max & + & - \\
\hdashline
ECB+ & N \& V & news & manual  & manual & manual & min & + & - \\
\hdashline
ECB+METAm & N \& V & news & manual*& semi-aut.* &  manual* & max & + & - \\
\hdashline
FCC-T & V & news (sport) & manual & manual & manual & min & - & - \\
\hdashline
GVC & V & news (violence) & manual & manual & manual & min & - & - \\
\hdashline
HyperCoref\textsubscript{exp} & V & news & autom.**  & autom. & autom. & max & - & + \\
\hdashline
MEANTIME\textsubscript{eng} & N \& V & news (economic) & manual & manual & manual & max & + & - \\
\hdashline
NewsWCL50r & N \& V & news (political) & manual & manual  & manual & max & - & + \\
\hdashline
NIdent\textsubscript{en-cd} & N & news (violence) & manual*  & manual & manual** & max & + & + \\
\hdashline
NP4E\textsubscript{cd} & N & news (violence) & manual & manual & semi-aut.** & max & + & - \\
\hdashline
WEC-Eng & V & news & autom.** & autom. &  autom. & min & - & - \\
\hline
\end{tabular}
\caption{The 12 selected publicly available datasets provide a balanced representation of event (V) and entity (N) coreference mentions. While uCDCR remains primarily focused on the news domain, it is supplemented with documents from three additional domains, such as emails and scientific literature. The length of annotated mention spans affects the lexical diversity of coreference chains and can influence the complexity of mention resolution. An asterisk (*) indicates that the source or annotation was reused from another dataset; for example, NIdent reannotates a document collection from NP4E, and ECB+METAm reuses document collection and chain labels from ECB+ while paraphrasing the annotated mentions. Two asterisks (**) indicate that the datasets required additional effort in preparation (see \Cref{sec:dataset_details}).}
\label{tab:annot}
\end{table*}

\section{Related work}
\label{sec:related}
CDCR research is generally framed as an event coreference resolution (ECR) task and predominantly relies on the ECB+ dataset \cite{cybulska-vossen-2014-using}, a state-of-the-art benchmark for cross-document event coreference. The ECB+ dataset primarily annotates event mentions, includes only a limited number of entity mentions—specifically when they function as event attributes (e.g., participants, locations, or dates), and establishes coreference links only when an event and all of its attributes are coreferential. 
Most CDCR models have been developed, trained, and evaluated exclusively on ECB+, including the most recent ones by 
\citet{hsu-horwood-2022-contrastive}, \citet{yu-etal-2022-pairwise}, \citet{ahmed-etal-2024-linear}, \citet{zhao-etal-2023-cross}, and \citet{nath-etal-2024-multimodal}. Only a few CDCR datasets have addressed the complexity of resolving entity mentions within this task, often employing dataset-specific methods. For instance, \citet{hamborg-2019-automated} introduced NewsWCL50, which includes annotated mentions susceptible to bias by word choice and labeling; \citet{zhukova-2022-xcoref} examined the resolution of these mentions with high lexical diversity. \citet{dakle-moldovan-2020-cerec} developed a model for resolving entity mentions, such as noun phrases, emails, and pronouns, from email threads. Similarly, \citet{ravenscroft-etal-2021-cd} proposed a model architecture tailored to the CD2CR dataset, which annotates mentions in a cross-domain setting. While dataset-tailored solutions can be effective for a specific task, they do not address model robustness across other datasets.

To mitigate model overfitting to the ECB+ dataset, recent research has increasingly focused on evaluating the generalizability of CDCR models across multiple datasets. \citet{bugert-2021-generalizing} were among the first to examine model performance in a cross-dataset setting, benchmarking existing approaches on three event-level CDCR datasets: ECB+ \cite{cybulska-vossen-2014-using}, GVC \cite{vossen-etal-2018-dont}, and FCC-T \citeplanguageresource{bugert-2021-generalizing}. They found that previous models performed well only on the datasets on which they were developed, trained, and evaluated, concluding that no single dataset can represent all domains or event types found in other datasets. Most subsequent research on generalizability has concentrated on ECR. For instance, \citet{eirew-etal-2021-wec} compared their model’s performance on ECB+ and their newly proposed WEC-Eng dataset, as did \citet{gao-etal-2024-enhancing}. \citet{bugert-gurevych-2021-event} explored improvements on ECB+, GVC, and FCC-T by augmenting them with the silver-quality HyperCoref dataset. \citet{ahmed-etal-2023-2} evaluated their approach using ECB+ and GVC, while \citet{ahmed-etal-2024-generating} compared model performance on ECB+ and its metaphor-reannotated variants, ECB+META1 and ECB+METAm. \citet{nath-etal-2024-multimodal} investigated a multimodal setup in event coreference detection using ECB+ and AIDA Phase 1 \citeplanguageresource{tracey-etal-2022-study}. Similarly, \citet{chen-etal-2025-employing} evaluated model performance across ECB+, WEC-Eng, and the metaphor-annotated ECB+META1 and ECB+METAm datasets. Notably, only \citet{held-etal-2021-focus} conducted a joint evaluation of both event and entity coreference, using ECB+, GVC, and FCC-T for events, and ECB+ together with CD2CR \cite{ravenscroft-etal-2021-cd} for entities.

To address the current fragmentation in CDCR research, we propose a unified dataset, \textbf{uCDCR}, that integrates event- and entity-level CDCR within a single framework. This dataset shifts the task from isolated mention-type resolution toward comprehensive general coreference resolution, unifies dataset formats to minimize preprocessing requirements, and promotes model generalizability in CDCR. We conduct systematic data analysis and evaluation of datasets, comparing their domain and topic coverage, annotation properties, and performance using established CDCR dataset comparison metrics.

\section{uCDCR dataset}
uCDCR consists of 12 datasets that jointly cover event and entity coreference resolution, as well as datasets that focus on either events or entities independently. This diversity in dataset design enables the study of CDCR without being constrained by a single common definition of ECR. Throughout the following sections, we will use the terms "coreference chains" and "coreference clusters" interchangeably.

In this section, we first introduce the CDCR datasets that collectively comprise uCDCR. \Cref{tab:annot} summarizes the qualitative characteristics of all datasets, including domain and topic composition, types of annotated mentions, sources of subtopic markers, annotation style for mentions and coreference chain identification, mention annotation span\footnote{Minimum span: \textit{"polygamist sect leader \underline{Warren Jeffs} was accused in"} vs. maximum span: \textit{"\underline{polygamist sect leader Warren Jeffs} was accused in"}}, and the types of identity relationships between mentions that form coreference chains. Second, we introduce a dataset unification framework for parsing and preprocessing that standardizes the format of all datasets and supplements any missing attributes on the dataset, mention, or coreference chain levels, which are required for comprehensive and systematic analysis. \Cref{sec:dataset_details} reports details on dataset unification and parsing.

\subsection{Datasets}
\label{sec:datasets}

The selected datasets for \textbf{uCDCR} are limited to those that are freely and publicly available and provide mention annotations at the phrase level.\footnote{The following CDCR datasets were excluded for the following reasons: Headlines of War \citeplanguageresource{radford-2020-seeing} provides annotations at the sentence level, while AIDA Phase 1 \citeplanguageresource{tracey-etal-2022-study}, EER \citeplanguageresource{hong-etal-2016-building}, cross-document Rich ERE \citeplanguageresource{song-etal-2018-cross}, and MCECR \citeplanguageresource{pouran-ben-veyseh-etal-2024-mcecr} are unavailable for download, either through direct links or via LDC.} 
\textbf{ECB+} \citeplanguageresource{cybulska-vossen-2014-using} extends EventCorefBank \citeplanguageresource{bejan-harabagiu-2010-unsupervised} by adding entity annotations and new subtopics, focusing primarily on event coreference. Its variant, \textbf{ECB+METAm} \citeplanguageresource{ahmed-etal-2024-generating}, enhances lexical diversity by applying GPT-4-based metaphoric paraphrasing to event mentions. 
\textbf{WEC-Eng} \citeplanguageresource{eirew-etal-2021-wec} and \textbf{FCC-T} \citeplanguageresource{bugert-2020-breaking}\footnote{We use a phrase-level version of the dataset proposed by \citetlanguageresource{bugert-2021-generalizing}.} address cross-subtopic event coreference. WEC-Eng extracts events from Wikipedia, using pivot pages as both subtopics and cluster names, while FCC-T focuses on football games and hyperlinks between events as coreference markers. 
\textbf{CD2CR} \citeplanguageresource{ravenscroft-etal-2021-cd} addresses cross-domain entity coreference by combining scientific and news articles to increase the lexical diversity of the coreference chains across domains. 
\textbf{GVC} \citeplanguageresource{vossen-etal-2018-dont} GVC targets gun violence events and annotates mentions and clusters them according to predefined event classes, such as "injury" or "missing". 
HyperCoref \citeplanguageresource{bugert-gurevych-2021-event} provides large-scale event-level CDCR from news hyperlinks in various topics. Since the dataset originally contains almost 3 million mentions, we parse only its experimental subset, \textbf{HyperCoref\textsubscript{exp}}, to make it comparable in size to the other datasets. 
NewsWCL50 \citeplanguageresource{hamborg-2019-automated} captures both event- and entity-level coreference, including bias-loaded phrases. The reannotated \textbf{NewsWCL50r}\citeplanguageresource{Zhukova2026b} contains more annotated mentions and chains but has more structured coreference relations. 
CEREC \citeplanguageresource{dakle-moldovan-2020-cerec} focuses on entity resolution in email threads, including semi-structured fields and cross-email coreference, with \textbf{CEREC\textsubscript{exp}} used in experiments. 
MEANTIME \citeplanguageresource{minard-etal-2016-meantime} provides a multilingual setup for event- and entity-level CDCR. Since the current version of uCDCR is English-only, we will use the \textbf{MEANTIME\textsubscript{eng}} version. 
NP4E \citeplanguageresource{hasler-etal-2006-nps} and NIdent \cite{recasens-etal-2010-typology}, \citeplanguageresource{recasens-etal-2012-annotating} are entity-focused datasets. NP4E annotates noun phrases of entities and those representing actions within terrorism-related subtopics with identity relations. NIdent-EN reannotates NP4E documents, capturing identity and near-identity relations. We use the cross-document versions of the datasets called \textbf{NP4E\textsubscript{cd}} and \textbf{NIdent\textsubscript{en-cd}}.

\begin{figure*}
    \centering
    \includegraphics[width=1\linewidth]{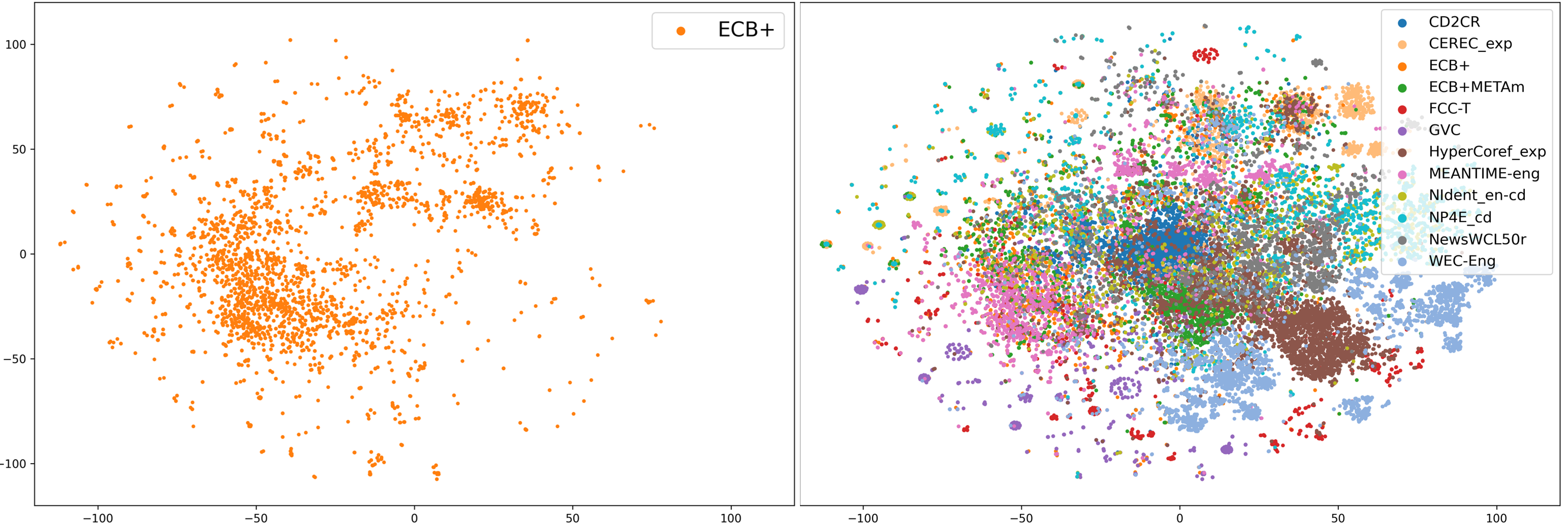}
    \caption{t-SNE representation of the mentions' texts first encoded with a sentence transformer. While ECB+ occupies 2/3 of the space, the combined uCDCR dataset covers the full space.  }
    \label{fig:ecb_vs_all}
\end{figure*}

\subsection{Methodology}
uCDCR contains all documents from their modifications as described in \Cref{sec:datasets}. To ensure the unified format of uCDCR, for each dataset, we underwent the following steps: (1) create a topic-subtopic-document structure, (2) collect (often) tokenized document texts and annotations, (3) re-parse the document texts to save document-token-document conversion information, (4) remap the annotated mentions to the re-parsed texts, (5) extract missing mention or chain attributes, (6) ensure the dataset splits, (7) create mention and document files.

\noindent\textbf{Create a topic-subtopic-document structure.} 
The CDCR datasets follow a three-level hierarchical structure comprising topics, subtopics, and documents. \textit{Topics} define the general subjects that link related documents.\footnote{For some datasets, such as GVC or FCC-T, a single topic was assigned, e.g., "football" to FCC-T.} \textit{Subtopics} divide a topic into collections of event-related documents, where a foundational event is typically characterized by four discourse elements \cite{hovy-etal-2013-events}: participant, action, location, and time.\footnote{In HyperCoref\textsubscript{exp} and WEC-Eng, we relied on the topic-subtopic structure based on the parsed hyperlinks or Wikipedia hierarchical topic structure; hence, this definition of subtopics does not hold.} \textit{Documents} describe individual events and may vary in type, including news articles, Wikipedia entries, or scientific publications. This hierarchical structure guides the application of coreference resolution, ranging from within-document to within-subtopic and cross-subtopic levels. CDCR is commonly framed in a within-subtopic setup; however, the recently introduced cross-subtopic CDCR presents a new challenge for models, primarily due to the increased number of mention candidates that must be resolved \citeplanguageresource{bugert-2020-breaking}.  

\noindent\textbf{Collect tokenized document texts and annotations.} 
The task of collecting the original data is tedious because the datasets use different input formats (e.g., XML-, JSON-, CSV-, or CoNLL-based), requiring careful tracking of the keys that link mentions to chains and documents, as well as the mention attributes and chain properties.

\noindent\textbf{Re-parse the document texts.} 
All datasets were originally tokenized using different tools and models, resulting in inconsistent parsing errors. For example, compound words like "hard-to-understand" could be tokenized as either a single token or as five separate tokens. To normalize tokenization across documents, we first concatenated tokens using a heuristic that accounts for whitespace following certain punctuation. Second, we corrected remaining whitespace issues using the whitespace correction method \cite{bast-etal-2023-fast}. Finally, the documents were re-parsed using spaCy (en\_core\_web\_sm), i.e., tokenized, dependency-parsed, POS-tagged, lemmatized, and NER-tagged, so that compound words appear as a single token and we can later extract missing mention-features. This re-parsing process also allowed us to preserve whitespace information, which can be used for re-annotation of documents or for visualizing annotated mentions and chains.

\noindent\textbf{Re-map the annotated mentions.} 
Remapping the annotated mentions ensures exact index alignment between mentions originally tokenized with different tools and the re-parsed documents.\footnote{Since whitespace correction and tokenization are using non-deterministic models, the models introduced certain errors in the text, resulting in 96.83-100\% of mentions compared to the original datasets (see \Cref{tab:parsing_rate}).} This step is essential for (a) accurate mapping to both the original documents and the context window provided for each mention, and (b) correct identification of the mention head and its associated attributes. Mentions are first remapped at the full-string level and, if unsuccessful, at the token-by-token level. 

\noindent\textbf{Extract missing mention or chain attributes.}
Following the format proposed by \citet{barhom-etal-2019-revisiting, eirew-etal-2021-wec}, we extract the following attributes for each mention based on its mapping to the spaCy-re-parsed documents: (1) the mention head (using dependency parsing) along with its lemma, POS tag, and NER label; (2) identifiers linking the mention to the original document, including topic ID, subtopic ID, document ID, sentence ID, and token IDs; and (3) the mention context, which maps the mention tokens to their surrounding context and aligns this context with the original document. The first type of attribute is necessary for data analysis, while the second is required for coreference resolution. The context is extracted according to a fair use or fair dealing policy,\footnote{"Fair use" in the United States and "fair dealing" in other jurisdictions permits the use of copyrighted material without permission under certain conditions, such as for purposes of commentary, criticism, or research.} as a window of $\pm$ 100 tokens, rounded to the nearest full sentence. We also generate chain-level attributes for each mention, such as its type (e.g., actor or location), when this information is not provided in the original annotation.
 
\noindent\textbf{Ensure the dataset splits.} 
For datasets with previously established splits, we reused the existing partitions. For several datasets, including MEANTIME and NP4E, we defined new splits. Most datasets were divided at the subtopic level into training, validation, and test sets, except for NewsWCL50r and ECB+METAm, which were originally designed as an evaluation-only dataset and were partitioned into validation and test sets.

\noindent\textbf{Create mention and document files.} 
The output for each dataset includes two separate JSON files per split: one containing event mentions and the other containing entity mentions. For datasets that provide full document texts,\footnote{HyperCoref\textsubscript{exp}, NewsWCL50r, and FCC-T will be released as mention-only files as their full texts were not originally publicly available.} we also release a Parquet file in a CoNLL-based format, i.e., with one token per line, preserving whitespaces from document re-parsing and references in the CoNLL format (i.e., with brackets). This structure with two JSON files supports the common approach of performing coreference resolution separately for events and entities, if needed.

\begin{table*}[]
\scriptsize
\centering
\begin{tabular}{|l|r|r|r|r|r|r|r|r|r|r|}
\hline
\multirow{2}{*}{\textbf{Dataset}} & \multirow{2}{*}{\textbf{Topics}} & \multirow{2}{*}{\textbf{Subtopics}} & \multirow{2}{*}{\textbf{Docs}} & \multirow{2}{*}{\makecell{\textbf{Tokens}}} & \multirow{2}{*}{\makecell{\textbf{Tokens}\\\textbf{/doc}}} & \multirow{2}{*}{\makecell{\textbf{Context}}} & \multirow{2}{*}{\textbf{Chains}} & \multirow{2}{*}{\textbf{Mentions}} & \multirow{2}{*}{\textbf{Singletons}} & \multirow{2}{*}{\makecell{\textbf{Mentions}\\\textbf{/doc}}} \\
& & & & & & & & & &  \\
\hline
CD2CR & 1 & 264 & 528 & 86K & 164 & 153 & 5222 & 7597 & 4496 & 14.4 \\
\hdashline
CEREC\textsubscript{exp} & 1 & 77 & 456 & 62K & 137 & 152 & 1475 & 7080 & 468 & 15.5 \\
\hdashline
ECB+ & 43 & 86 & 976 & 628K & 643 & 149 & 4952 & 15051 & 3445 & 15.4 \\
\hdashline
ECB+METAm & 18 & 36 & 402 & 184K & 459 & 161 & 2095 & 6348 & 1481 & 15.8 \\
\hdashline
FCC-T & 1 & 183 & 428 & 355K & 829 & 214 & 469 & 3561 & 254 & 8.3 \\
\hdashline
GVC & 1 & 241 & 510 & 185K & 364 & 183 & 1679 & 7284 & 635 & 14.3 \\
\hdashline
HyperCoref\textsubscript{exp} & 35 & 324 & 40938 & 29M & 727 & 184 & 13102 & 60401 & 5869 & 1.5 \\
\hdashline
MEANTIME\textsubscript{eng} & 4 & 120 & 120 & 53K & 442 & 181 & 2938 & 6506 & 2270 & 54.2 \\
\hdashline
NewsWCL50r & 10 & 10 & 50 & 50K & 992 & 223 & 433 & 6531 & 102 & 130.6 \\
\hdashline
NIdent\textsubscript{en-cd} & 1 & 5 & 93 & 50K & 541 & 212 & 2463 & 12988 & 1275 & 139.7 \\
\hdashline
NP4E\textsubscript{cd} & 1 & 5 & 94 & 51K & 545 & 209 & 667 & 6559 & 0 & 69.8 \\
\hdashline
WEC-Eng & 18 & 7370 & 37129 & 5054K & 136 & 149 & 7597 & 43672 & 865 & 1.2 \\
\hline
\textbf{uCDCR} & 115 & 8680 & 81229 & 36M & 498* & 181* & 43089 & 183578 & 21160 & 40.1* \\
\hline
\end{tabular}
\caption{General statistics of the datasets in uCDCR. We do not include NIdent\textsubscript{en-cd} and ECB+METAm in the uCDCR statistics on topics, subtopics, documents, and tokens because they reannotate documents from other datasets. An asterisk (*) indicates an average value.}
\label{tab:general_stats}
\end{table*}

\begin{table*}[]
\scriptsize
\centering
\begin{tabular}{|l||c|c||c|c||c|c||c|c||c|c|}
\hline
\multirow{2}{*}{\textbf{Dataset}} & \multicolumn{2}{c||}{\textbf{Cross-subtopic}} & \multicolumn{2}{c||}{\textbf{Chains}} & \multicolumn{2}{c||}{\textbf{Mentions}} & \multicolumn{2}{c||}{\textbf{Singletons}} & \multicolumn{2}{c|}{\textbf{Average chain size}}  \\
\cline{2-11}
& chains & mentions & entities & events & entities & events & entities & events  & \makecell{with singl.} & \makecell{w/o singl.} \\
\hline
CD2CR & 0 & 0 & 5222 & 0 & 7597 & 0 & 4496 & 0 & 1.45 & 4.27 \\
\hdashline
CEREC\textsubscript{exp} & 0 & 0 & 1475 & 0 & 7080 & 0 & 468 & 0 & 4.80 & 6.67 \\
\hdashline
ECB+ & 67 & 1291 & 4016 & 936 & 10052 & 4999 & 3211 & 234 & 3.04 & 7.70 \\
\hdashline
ECB+METAm & 34 & 700 & 1711 & 384 & 4166 & 2182 & 1401 & 80 & 3.03 & 7.93 \\
\hdashline
FCC-T & 252 & 3302 & 0 & 469 & 0 & 3561 & 0 & 254 & 7.59 & 11.89 \\
\hdashline
GVC & 0 & 0 & 0 & 1679 & 0 & 7284 & 0 & 635 & 4.34 & 6.37 \\
\hdashline
HyperCoref\textsubscript{exp} & 3287 & 26283 & 4280 & 8822 & 7300 & 53101 & 3222 & 2647 & 4.61 & 7.15 \\
\hdashline MEANTIME\textsubscript{eng} & 164 & 2835 & 816 & 2122 & 3855 & 2651 & 487 & 1783 & 2.21 & 6.22 \\
\hdashline
NewsWCL50r & 0 & 0 & 374 & 59 & 4758 & 1773 & 102 & 0 & 15.08 & 19.42 \\
\hdashline
NIdent\textsubscript{en-cd} & 0 & 0 & 2463 & 0 & 12988 & 0 & 1275 & 0 & 5.27 & 9.86 \\
\hdashline
NP4E\textsubscript{cd}& 0 & 0 & 664 & 0 & 6559 & 0 & 0 & 0 & 9.88 & 9.88 \\
\hdashline
WEC-Eng & 2501 & 22359 & 0 & 7597 & 0 & 43672 & 0 & 1032 & 5.75 & 6.36 \\
\hline
\textbf{uCDCR} & 6305 & 56770 & 21021 & 22068 & 64355 & 119223 & 14662 & 6665 & 5.59 & 8.64 \\
\hline
\end{tabular}
\caption{The overview of the datasets highlights their distribution across subtopics and their composition in terms of events and entities. A few event-centric datasets specifically tackle the challenge of annotating and resolving mentions across subtopics, aiming to capture the complexity of event characteristics. Also, only a few datasets address both entity and event resolution jointly, while most focus exclusively on one type of mention.}
\label{tab:mention}
\end{table*}

\begin{figure}[h]
    \centering
    \includegraphics[width=1\linewidth]{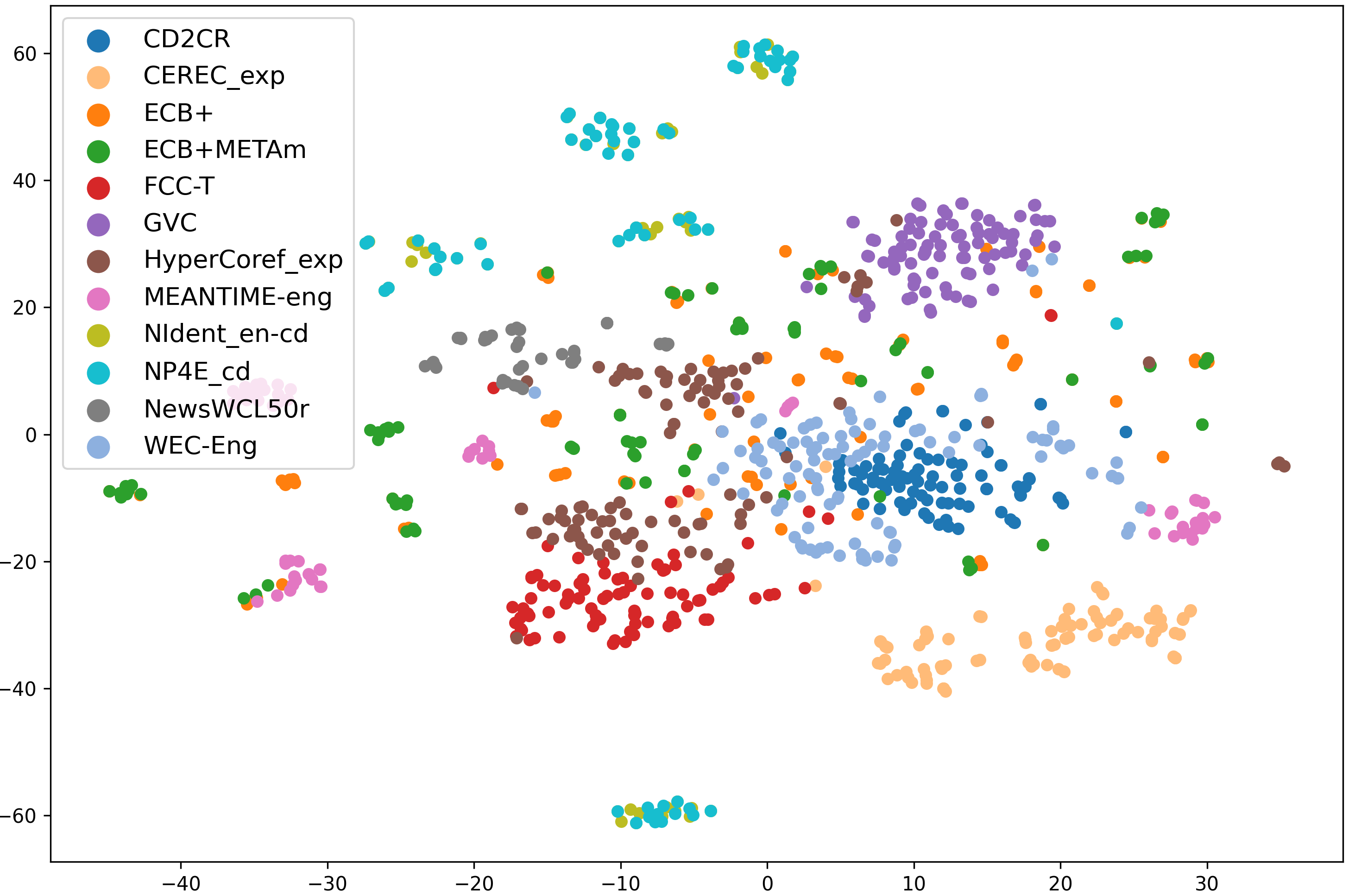}
    \caption{t-SNE representations for 100 randomly selected documents from each dataset. While most datasets cluster closely by topic, some are positioned farther apart, thereby increasing overall diversity in uCDCR.}
    \label{fig:doc_scatterplot}
\end{figure}

\section{Quantitative analysis}
\label{sec:quant}
This section presents a comprehensive analysis of all datasets comprised in uCDCR. We first present general statistics describing the overall dataset composition, including the numbers of subtopics, documents, mentions, and coreference chains across sources. Next, we analyze the datasets with respect to lexical diversity and ambiguity, focusing on how variation in lexical expression and form affects the difficulty of cross-document coreference resolution. Finally, we evaluate the dataset performance using a lemma-based baseline, which serves as a reference point for understanding how lexical variation influences model accuracy. 

Since ECB+ serves as the current benchmark dataset for CDCR, most of our comparisons will be against it to highlight the need for a more diverse set of datasets for evaluating CDCR models.

\subsection{General statistics}
As a general overview of the datasets, we report key statistics for each dataset, including the number of topics, subtopics, documents, chains, mentions, and singleton clusters (i.e., clusters containing only a single mention). Additionally, we report the average document size (i.e., the number of tokens per document) and annotated mention density (i.e., the number of mentions per document). \Cref{tab:general_stats} shows that HyperCoref\textsubscript{exp} and WEC-Eng are the largest datasets in terms of subtopics, documents, annotated mentions, and coreference chains. Although NIdent\textsubscript{en-cd} and NewsWCL50r contain the fewest documents, they are the most densely annotated datasets, each containing approximately 10 times as many annotated mentions per document as ECB+. The low annotation density in HyperCoref\textsubscript{exp} and WEC-Eng reflects their construction methodology, which relies on crawling event hyperlinks in news or Wiki articles to create coreference chains (one link per pair of documents). The number of singletons suggests that these mentions can serve as a valuable source for generating negative mention pairs during training a cross-encoder model \cite{cattan-etal-2021-realistic, caciularu-etal-2021-cdlm-cross}, though they should be handled cautiously and excluded from evaluation to avoid bias of the inflated metrics \citep{cattan-etal-2021-realistic}.

To illustrate the distribution of document topics, we randomly selected up to 100 documents from each dataset using a uniform distribution to ensure equal representation, and projected them into t-SNE space, as shown in \Cref{fig:doc_scatterplot}. We see that ECB+/ECB+METAm exhibit sparse, center-based topic distributions, whereas the other datasets form more consolidated clusters. Notably, NP4E\textsubscript{cd}/NIdent\textsubscript{en-cd} and MEANTIME\textsubscript{eng} form distinct clusters, corresponding to their narrowly-defined event-specific subtopics. Similarly, the majority of NewsWCL50r documents belong to a single cluster, likely due to frequent reporting on US politics, and we see that NewsWCL50r overlaps with the part of HyperCoref\textsubscript{exp} also related to the US politics. The other part of HyperCoref\textsubscript{exp} related to the sport topics overlaps with FCC-T. In contrast, CD2CR and CEREC\textsubscript{exp} display distinct clusters related to science and email correspondence. Similarly to documents, we randomly selected 6300 mentions and projected them in t-SNE space.\footnote{Each mention text is first encoded with a text encoder and then projected into a t-SNE space.} \Cref{fig:ecb_vs_all} shows that while ECB+ occupies 2/3 of the space, the combined datasets cover the space almost fully, hence allowing for topic and lexical variation.

\Cref{tab:mention} shows that the number of event and entity coreference chains becomes almost equal when combined across 12 datasets, even though the actual ratio of entity to event mentions is approximately 1:2. Excluding singleton mentions further reduces this proportion, for example, because ECB+ often annotates entities as event attributes linked to specific event occurrences and not always liked as coreferences. Moreover, \Cref{tab:mention} illustrates that singletons significantly influence average chain sizes, i.e., the number of mentions per chain. NewsWCL50r and FCC-T contain the largest chains, which may be challenging to fully resolve, as more mentions must be correctly assigned to a single chain.

\begin{table}[h]
\scriptsize
\centering
\begin{tabular}{|l|c|c|c|}
\hline
\multirow{2}{*}{\textbf{Dataset}} &  \multicolumn{3}{c|}{\textbf{Dataset split}} \\
\cline{2-4}
 & train & val & test \\
\hline
CD2CR & 4602 / 3221 & 1820 / 1245 & 1175 / 756 \\
\hdashline
CEREC\textsubscript{exp}  & 3719 / 750 & 2094 / 422 & 1267 / 303 \\
\hdashline
ECB+ & 8521 / 2810  & 2710 / 737 & 3820 / 1408 \\
\hdashline
ECB+METAm & 0 / 0 & 2619 / 719 & 3729 / 1376 \\
\hdashline
FCC-T  & 1603 / 259 & 750 / 136 & 1208 / 191 \\
\hdashline
GVC & 5299 / 1158 & 977 / 271 & 1008 / 250 \\
\hdashline
HyperCoref\textsubscript{exp} &  49777 / 5477 & 5865 / 3904 & 4759 / 3721 \\
\hdashline
MEANTIME\textsubscript{eng} &  3355 / 1522 & 1895 / 756 & 1256 / 686 \\
\hdashline
NewsWCL50r & 0 / 0 & 2740 / 145 & 3791 / 686  \\
\hdashline
NIdent\textsubscript{en-cd} & 7178 / 1316 & 3180 / 800 & 2630 / 347 \\
\hdashline
NP4E\textsubscript{cd} & 3375 / 332 & 1744 / 199 & 1440 / 136 \\
\hdashline
WEC-Eng  & 40529 / 7042 & 1250 / 233 & 1893 / 322 \\
\hline
\textbf{uCDCR} & 128K / 24K & 27644 / 9567 & 27976 / 9783 \\
\hline
\end{tabular}
\caption{The dataset composition of uCDCR split into train, validation, and test sets (in \textit{mentions / chains}). ECB+METAm and NewsWCL50r are for validation and testing only.}
\label{tab:uCDCR}
\end{table}

Finally, \Cref{tab:uCDCR} shows the composition of the train, validation, and test splits of uCDCR. The training split is highly unbalanced, with some datasets (e.g., WEC-Eng, HyperCoref\textsubscript{exp}) contributing 40K–50K mentions, while others (e.g., ECB+METAm, NewsWCL50r) appear only in validation and test. By contrast, the test split is more balanced, with all datasets containing 1,200–4,700 mentions, allowing for a fair evaluation.

\subsection{Lexical diversity and ambiguity}
Lexical diversity and lexical ambiguity are key linguistic properties that influence the difficulty of cross-document coreference resolution.  

\textbf{Lexical diversity} captures the extent of variation in surface forms used to refer to the same entity or event. It is quantified using three complementary measures: (1) the average number of unique head lemmas per cluster (UL-a) \cite{eirew-etal-2021-wec}, which reflects the lexical variability within individual coreference chains; (2) the overall number of unique head lemmas across the dataset (UL-o) \cite{bugert-gurevych-2021-event}, indicating the global lexical richness of mentions; (3) the phrasing diversity (PD) \cite{zhukova-etal-2022-towards}, which measures the degree of paraphrastic variation among coreferential mentions, and (4) MTLD (Measure of Textual Lexical Diversity) \citep{McCarthy2010, ahmed-etal-2024-generating}, which evaluates lexical diversity as the range of unique words used in a text, i.e., concatenated mentions' tokens of one chain, while minimizing sensitivity to text length. Although most lexical metrics are lemma-driven, the MTLD metric captures a different linguistic aspect. We calculate PD and MLTD as weighted averages of the chain sizes. All metrics are computed on non-singleton chains. Because the UL-o metric counts the absolute number of unique lemmas, we computed it exclusively on the test set, which has the most balanced split across datasets (see \Cref{tab:uCDCR}), whereas all other metrics are computed over the entire dataset. 

\textbf{Lexical ambiguity}, in contrast, quantifies how often the same lexical form refers to different entities or events. It is measured by the average number of clusters sharing the same head lemma (AL) \cite{eirew-etal-2021-wec}, which reflects the ambiguity of a given lexical form across contexts. AL is also computed on the non-singletons of the full datasets.

\begin{table}[h]
\centering
\scriptsize
\begin{tabular}{|l|r|r|r|r|r|}
\hline
\multirow{2}{*}{\textbf{Dataset}}  &  \makecell{\textbf{Lexical}} & \multicolumn{4}{c|}{\textbf{Lexical diversity}}  \\
\cline{3-6}
  & \textbf{ambiguity} & UL-a & UL-o* & PD & MTLD  \\
\hline
CD2CR  & 1.97 & 3.21 & 295 & \textbf{\textit{5.94}} & 10.50  \\
\hdashline
CEREC\textsubscript{exp} & 2.25 & 3.07 & 306 & 0.94 & 4.67 \\
\hdashline
ECB+ & 2.22 & 2.61 & 605 & 1.50 & 4.52 \\
\hdashline
ECB+METAm & 1.9 & 4.22 & \textbf{894} & 4.71 & 8.78 \\
\hdashline
FCC-T & 4.32 & \underline{\textit{4.90}} & 257 & \underline{\textit{5.02}} & 6.11 \\
\hdashline
GVC & \textbf{16.48} & 3.09 & 89 & 1.94 & 3.77 \\
\hdashline
HyperCoref\textsubscript{exp} & \underline{\textit{5.75}} & \textbf{\textit{5.47}} & \textbf{\textit{846}} & \textbf{9.07} & \textbf{16.70} \\
\hdashline
MEANTIME\textsubscript{eng} & 2.12 & 2.31 & 177 & 2.10 & 6.24 \\
\hdashline
NewsWCL50r & 1.89 & \textbf{6.05} & \underline{\textit{814}} & \textbf{9.06} & \textbf{\textit{15.58}} \\
\hdashline
NIdent\textsubscript{en-cd}& 2.57 & 2.68 & 317 & 3.03 &  \underline{\textit{12.63}} \\
\hdashline
NP4E\textsubscript{cd} & 2.54 & 2.72 & 174 & 1.58 & 10.59 \\
\hdashline
WEC-Eng & \textbf{\textit{6.12}} & 1.96 & 229 & 2.25 & 6.95 \\
\hline
\end{tabular}
\caption{Analysis of coreference mention properties: lexical diversity and lexical ambiguity. Lexical diversity reflects the difficulty of linking mentions that express the same concept using different phrasings, whereas lexical ambiguity captures the difficulty of distinguishing between mentions that are lexically similar but refer to different entities or events. An asterisk (*) indicates a value computed on a test set.}
\label{tab:lexical}
\end{table}

\begin{table}[h]
\scriptsize
\begin{tabular}{|l|c|c|c|c|c|}
\hline
\textbf{Dataset} & \textbf{Level} & \textbf{MUC} & \textbf{B3} & \textbf{CEAF\textsubscript{e}} & \textbf{CoNLL} \\
\hline
CD2CR & \multirow{13}{*}{\rotatebox{90}{subtopic}} & 39.62 & 54.26 & 33.39 & 42.42 \\
\hdashline
CEREC\textsubscript{exp} &   & 69.57 & 53.88 & 37.20 & 53.55 \\
\hdashline
ECB+ &   & 78.38 & 62.10 & 42.33 & 60.94 \\
\hdashline
ECB+METAm &  & 62.36 & 50.62 & 28.25 & 47.08 \\
\hdashline
FCC-T &   & 57.55 & 73.88 & 62.56 & \underline{\textit{64.66}} \\
\hdashline
GVC &  & 62.04 & 57.83 & 37.87 & 52.58 \\
\hdashline
HyperCoref\textsubscript{exp} &  & 13.25 & 69.90 & 53.66 & 45.61 \\
\hdashline
MEANTIME\textsubscript{eng} &  & 56.87 & 74.79 & 65.54 & \textbf{\textit{65.73}} \\
\hdashline
NewsWCL50r &   & 79.75 & 50.35 & 23.09 & 51.06 \\
\hdashline
NIdent\textsubscript{en-cd} &  & 84.25 & 46.34 & 39.80 & 56.80 \\
\hdashline
NP4E\textsubscript{cd} &  & 84.16 & 51.82 & 44.57 & 60.18 \\
\hdashline
WEC-Eng &    & 87.66 & 86.45 & 74.95 & \textbf{83.02} \\
\hline
\textbf{uCDCR} &  & 64.62 & 61.02 & 45.27 & 56.97 \\
\hline
ECB+ & \multirow{7}{*}{\rotatebox{90}{topic}}  & 78.41 & 58.83 & 41.68 & \textbf{59.64} \\
\hdashline
ECB+METAm &  & 62.99 & 48.50 & 27.49 & 46.33 \\
\hdashline
FCC-T &  & 63.61 & 39.71 & 19.63 & 40.98 \\
\hdashline
HyperCoref\textsubscript{exp} & &  13.94 & 66.09 & 48.38 & 42.80 \\
\hdashline
MEANTIME\textsubscript{eng} &  & 72.28 & 54.12 & 46.49 & \underline{\textit{57.63}} \\
\hdashline
WEC-Eng &   & 83.38 & 53.52 & 38.73 & \textbf{\textit{58.54}} \\
\hline
\textbf{uCDCR} &  & 62.43 & 53.46 & 37.07 & 50.99\\
\hline
\end{tabular}
\caption{Performance analysis of the datasets using the same-head-lemma baseline on the subtopic and topic levels.  }
\label{tab:topic-subtopic}
\end{table}

\Cref{tab:lexical} summarizes the lexical properties of the CDCR datasets, showing that GVC is the most lexically ambiguous dataset, whereas HyperCoref\textsubscript{exp} and NewsWCL50r are the most lexically diverse according to the three combined metrics. ECB+ has the second lowest lexical diversity measured by PD and MLTD. Our metric reproduction aligned with the values and trends reported upon the introduction of these metrics. The values for AL and UL-a align with those reported by \citet{eirew-etal-2021-wec} for WEC-Eng, MEANTIME\textsubscript{eng}, ECB+, and GVC, confirming the reproducibility of our comparison. We also replicated the magnitude of UL-o scores reported by \citet{bugert-gurevych-2021-event} for ECB+, FCC-T, GVC, and HyperCoref\textsubscript{exp}, as well as the PD scores from \citet{zhukova-etal-2022-towards} for NewsWCL50r and ECB+. When using MTLD, we reproduced the general trend observed in \citet{ahmed-etal-2024-generating}, showing that ECB+METAm is more lexically diverse than ECB+. However, the exact magnitudes differed, as \citet{ahmed-etal-2024-generating} did not provide implementation details for MTLD. Consequently, we implemented the metric ourselves following the original paper and applied it to the joined entity and event mentions rather than to event mentions alone.

Analyzing the most lexically diverse datasets across all metrics reveals several key properties that contribute to their diversity: (1) Annotation of mentions from a large number of sources, as texts authored by many individuals exhibit varied writing styles and potential polarity drift; (2) Annotation of coreference chains spanning multiple domains, which increases linguistic and stylistic variation, e.g., news vs. science; (3) Inclusion of figurative language, metaphors, euphemisms, and other forms of “looser” coreference, which are often context-specific and used to frame events or entities in particular ways; and (4) Annotation of the same event from multiple perspectives, for instance, a football match being described as a success or a failure depending on which team won.

\subsection{Baseline performance analysis}
Although the lexical diversity metric provides valuable insights into the complexity of CDCR across datasets, the most widely used baseline for performance analysis remains the same-head-lemma baseline. In this approach, mentions that share the same lemma of their head phrase are grouped together and compared to the gold chains. We calculate this baseline on the test split using only non-singleton chains to avoid performance inflation caused by singletons \citep{cattan-etal-2021-realistic}. Furthermore, we compute metrics at two levels: the subtopic level, which serves as the default for all datasets, and the topic level for datasets containing cross-subtopic coreference chains (see \Cref{tab:mention}). Lastly, we compute the metrics after resolving events and entities separately. The reported metrics are averages across subtopics or topics within each dataset.

We evaluate model performance using the state-of-the-art coreference resolution metrics, i.e., MUC \cite{vilain-etal-1995-model}, which measures link-based recall and precision between predicted and gold clusters; B3 \cite{bagga-baldwin-1998-entity}, which computes mention-level precision and recall; and CEAF\textsubscript{e} \cite{luo-2005-coreference}, which aligns predicted and gold clusters based on entity similarity. We also report the F1-CoNLL score, calculated as the harmonic mean of the three F1 scores (MUC, B3, and CEAF\textsubscript{e}), following the CoNLL shared task convention \cite{pradhan-etal-2014-scoring}. We use the official evaluation scripts released by \citet{pradhan-etal-2014-scoring} for MUC, B3, CEAF\textsubscript{e}, and F1-CoNLL. We report only F1-scores and later report all scores in \Cref{tab:full_performance}.

\Cref{tab:topic-subtopic} presents the evaluation results of the same-lemma baseline and shows that WEC-Eng is the most easily solvable dataset under this baseline at the subtopic level, followed by MEANTIME\textsubscript{eng}, and FCC-T. Although WEC-Eng is the second-highest in lexical ambiguity, it also has one of the lowest lexical diversities, making the dataset "solvable" even with a simple baseline. The baseline performs better on ECB+ than on average across the dataset, both on the subtopic and topic level (60.94 vs. 56.97 and 59.64 vs. 50.99), indicating that although ECB+ has a good average representation of uCDCR across topics, mentions, and performance coverage, CDCR models should be evaluated on more diverse, complex, or simpler-to-resolve datasets. The most challenging datasets for this baseline are CD2CR, HyperCoref\textsubscript{exp}, ECB+METAm, and NewsWCL50r on a subtopic level, and FCC-T and HyperCoref\textsubscript{exp} at the topic level. In general, these values correlate with one of the high values of lexical diversity metrics but not with the highest value of lexical ambiguity. Yet the resolution of GVC's mentions, which has the highest lexical ambiguity, performs worse than average on the other datasets, indicating that mentions disambiguation is a challenging task.

\begin{table}[h]
\centering
\scriptsize
\begin{tabular}{|l|c|c|c|c|c|}
\hline
\textbf{Dataset} & \textbf{Type} & \textbf{MUC} & \textbf{B3} & \textbf{CEAF\textsubscript{e}} & \textbf{CoNLL} \\
\hline
CD2CR & \multirow{10}{*}{\rotatebox{90}{entity}} & 39.62 & 54.26 & 33.39 & 42.42 \\
\hdashline
CEREC\textsubscript{exp} &  & 69.57 & 53.88 & 37.20 & 53.55 \\
\hdashline
ECB+ &  & 78.09 & 60.53 & 40.71 & \underline{\textit{59.77}} \\
\hdashline
ECB+METAm & & 77.86 & 60.04 & 40.67 & \underline{\textit{59.52}} \\
\hdashline
HyperCoref\textsubscript{exp} &  & 17.75 & 76.27 & 60.60 & 51.54 \\
\hdashline
MEANTIME\textsubscript{eng} &  & 57.17 & 73.67 & 65.12 & \textbf{65.32} \\
\hdashline
NewsWCL50r &  & 84.66 & 56.57 & 33.13 & 58.12 \\
\hdashline
NIdent\textsubscript{en-cd} & & 84.25 & 46.34 & 39.80 & 56.80 \\
\hdashline
NP4E\textsubscript{cd} &   & 84.16 & 51.82 & 44.57 & \textbf{60.18} \\
\hdashline
\hline
\textbf{uCDCR} &  & 65.90 & 59.26 & 43.91 & 56.36 \\
\hline
ECB+ & \multirow{9}{*}{\rotatebox{90}{event}} & 49.76 & 71.02 & 42.81 & 61.10 \\
\hdashline
ECB+METAm &  & 21.51 & 49.98 & 18.06 & 28.26 \\
\hdashline
FCC-T &  & 69.91 & 68.64 & 62.56 & \underline{\textit{64.66}} \\
\hdashline
GVC &  & 44.61 & 64.29 & 37.87 & 52.58 \\
\hdashline
HyperCoref\textsubscript{exp} &   & 52.70 & 68.25 & 48.84 & 43.00 \\
\hdashline
MEANTIME\textsubscript{eng} & & 70.56 & 79.89 & 67.54 & \textbf{\textit{68.65}} \\
\hdashline
NewsWCL50r &   & 14.60 & 38.62 & 4.46  & 30.08 \\
\hdashline
WEC-Eng &  & 80.09 & 90.02 & 74.95 & \textbf{83.02} \\
\hline
\textbf{uCDCR} &  & 50.47 & 66.34 & 44.63 & 53.92 \\
\hline
\end{tabular}
\caption{Performance analysis of the datasets using the same-head-lemma baseline separately for events and entities on the subtopic level. Although CDCR is often presented as ECR, resolving entity mentions is comparably complex to resolving events when several datasets are considered.}
\label{tab:performance_by_type}
\end{table}

\Cref{tab:performance_by_type} compares the baseline performance when applied separately to event and entity mentions at the subtopic level and shows that the performance metric is comparable between these types (53.92 vs. 56.36). 
Therefore, excluding entity resolution from CDCR as a standalone task simplifies it and reduces model generalizability. Additionally, we observed a direct impact of the annotation guidelines on performance analysis: when a dataset focuses on increasing the variety of phrasing in event mentions, as in ECB+METAm, or on annotating more complex near-identity relations, as in NewsWCL50r, the baseline's performance drops significantly (28.29 and 30.08, correspondingly). Thus, the more diverse specific mentions or chain properties each coding book annotates, the more exposed a CDCR model becomes to linguistic nuances and the more robust it becomes.

\subsection{Discussion and future work}
Despite significant progress in developing individual CDCR datasets, the absence of a unified benchmark has long hindered systematic advancement in the field. Existing datasets vary substantially in their domain coverage, ranging from encyclopedic or scientific sources to news-based corpora, and structural formats, making direct comparison across studies difficult. For example, the datasets integrated into uCDCR differ not only in the types of mentions and coreference chains they annotate, i.e., events and/or entities, but also in their annotation guidelines and stylistic diversity. These variations produce markedly different levels of lexical diversity and ambiguity, which in turn influence baseline and model performance, as shown in our same-lemma analysis (\Cref{tab:topic-subtopic}).

The creation of uCDCR addresses this fragmentation by providing a unified framework that harmonizes data formats, evaluation protocols, and lexical diversity metrics across datasets. Such standardization enables consistent benchmarking and encourages a more reliable assessment of model generalization across domains and annotation paradigms. Similar unification efforts in other NLP subfields, such as GLUE \citeplanguageresource{wang-etal-2018-glue} and SuperGLUE \citeplanguageresource{Wang2019} for sentence understanding, or BIG-Bench for large language model evaluation \citeplanguageresource{kazemi-etal-2025-big}, have catalyzed rapid methodological progress by providing a shared experimental foundation and transparent evaluation standards. 
In future work, we plan to systematically evaluate how state-of-the-art CDCR models perform across the compiled datasets in order to assess their generalisability and cross-dataset robustness.

\section{Conclusion}
In this work, we introduced uCDCR, the first unified benchmark for CDCR, which integrates diverse publicly available datasets spanning both entity- and event-level across different annotation guidelines, topics, and domains. By standardizing formats, correcting tokenization inconsistencies, and completing missing attributes, uCDCR provides a framework for reproducible evaluation across originally heterogeneous datasets. Our framework covers all available metrics for dataset descriptions, and the systematic analysis reveals how lexical diversity, ambiguity, and annotation strategies vary across datasets and how they impact baseline performance, underscoring the importance of dataset diversity for robust model development.

\section*{Limitations}
While uCDCR represents a significant step toward standardizing CDCR, several limitations remain. 

First, the benchmark currently focuses exclusively on English datasets, which restricts its applicability to multilingual or cross-lingual scenarios involving CDCR. 

Second, although uCDCR integrates diverse domains and topics, some corpora remain underrepresented, such as social media, informal communication, or emerging news sources, which may limit generalization to these genres. 

Third, the quality and granularity of annotations vary across the included datasets, reflecting differences in the original annotation guidelines. Despite harmonization efforts, residual inconsistencies may affect cross-dataset comparability. Although we aimed to include as many mention- and chain-level attributes as possible, identifying them in one dataset and extracting or generating them for others, some dataset-specific information may still be missing. We acknowledge that the absence of this information may introduce inconsistencies in cross-dataset comparisons and analyses. Moreover, we do not report details about the annotation guidelines, annotation process, or intercoder agreement. To mitigate this limitation, we release the parsing code for all datasets, encouraging the extraction of more information from the dataset and enabling more detailed analysis within our proposed unified framework. 

Fourth, while the same-head-lemma baseline provides a useful reference point, it does not capture more complex linguistic phenomena, such as metaphorical or context-dependent coreference, which remain challenging for current models. Benchmarking published CDCR models is required to provide a complete comparison of their performance on these datasets. 

Fifth, uCDCR’s construction relies on publicly available datasets, meaning it cannot fully address the bias, under-annotation, or topic imbalance inherent in the original corpora. 

Finally, some previously used CDCR datasets, such as AIDA Phrase 1, could not be included due to unavailability. Moreover, datasets distributed under LDC licenses typically impose restrictions preventing modification and redistribution, unlike those released under share-alike policies.

\section*{Acknowledgments}
We thank Jonas Becker for helping with dataset parsing, Michael Bugert for helping navigate HyperCoref, Zhuojing Huang for annotating and validating the datasets, and Sergio Gonzalez Orjuela for testing the datasets during CDCR model development.

\section{Bibliographical References}
\label{sec:reference}
\bibliographystyle{lrec2026-natbib}
\bibliography{lrec2026-example}

\begin{thebibliography}{23}
\expandafter\ifx\csname natexlab\endcsname\relax\def\natexlab#1{#1}\fi

\bibitem[{Ahmed et~al.(2024)Ahmed, Wang, Baker, Stowe, and Martin}]{ahmed-etal-2024-generating}
Shafiuddin~Rehan Ahmed, Zhiyong~Eric Wang, George~Arthur Baker, Kevin Stowe, and James~H. Martin. 2024.
\newblock \href {https://doi.org/10.18653/v1/2024.acl-short.27} {Generating harder cross-document event coreference resolution datasets using metaphoric paraphrasing}.
\newblock In \emph{Proceedings of the 62nd Annual Meeting of the Association for Computational Linguistics (Volume 2: Short Papers)}, pages 276--286, Bangkok, Thailand. Association for Computational Linguistics.

\bibitem[{Bejan and Harabagiu(2010)}]{bejan-harabagiu-2010-unsupervised}
Cosmin Bejan and Sanda Harabagiu. 2010.
\newblock \href {https://aclanthology.org/P10-1143/} {Unsupervised event coreference resolution with rich linguistic features}.
\newblock In \emph{Proceedings of the 48th Annual Meeting of the Association for Computational Linguistics}, pages 1412--1422, Uppsala, Sweden. Association for Computational Linguistics.

\bibitem[{Bugert and Gurevych(2021)}]{bugert-gurevych-2021-event}
Michael Bugert and Iryna Gurevych. 2021.
\newblock \href {https://doi.org/10.18653/v1/2021.emnlp-main.38} {{E}vent coreference data (almost) for free: {M}ining hyperlinks from online news}.
\newblock In \emph{Proceedings of the 2021 Conference on Empirical Methods in Natural Language Processing}, pages 471--491, Online and Punta Cana, Dominican Republic. Association for Computational Linguistics.

\bibitem[{Bugert et~al.(2020)Bugert, Reimers, Barhom, Dagan, and Gurevych}]{bugert-2020-breaking}
Michael Bugert, Nils Reimers, Shany Barhom, Ido Dagan, and Iryna Gurevych. 2020.
\newblock \href {https://ceur-ws.org/Vol-2593/paper3.pdf} {Breaking the subtopic barrier in cross-document event coreference resolution}.
\newblock In \emph{Proceedings of Text2Story — Third Workshop on Narrative Extraction From Texts co-located with 42nd European Conference on Information Retrieval (ECIR 2020), Virtual Event}, Lisbon, Portugal. CEUR.

\bibitem[{Bugert et~al.(2021)Bugert, Reimers, and Gurevych}]{bugert-2021-generalizing}
Michael Bugert, Nils Reimers, and Iryna Gurevych. 2021.
\newblock \href {https://doi.org/10.1162/coli_a_00407} {{Generalizing Cross-Document Event Coreference Resolution Across Multiple Corpora}}.
\newblock \emph{Computational Linguistics}, 47(3):575--614.

\bibitem[{Cybulska and Vossen(2014)}]{cybulska-vossen-2014-using}
Agata Cybulska and Piek Vossen. 2014.
\newblock \href {https://aclanthology.org/L14-1646/} {Using a sledgehammer to crack a nut? lexical diversity and event coreference resolution}.
\newblock In \emph{Proceedings of the Ninth International Conference on Language Resources and Evaluation ({LREC}'14)}, pages 4545--4552, Reykjavik, Iceland. European Language Resources Association (ELRA).

\bibitem[{Dakle and Moldovan(2020)}]{dakle-moldovan-2020-cerec}
Parag~Pravin Dakle and Dan Moldovan. 2020.
\newblock \href {https://doi.org/10.18653/v1/2020.coling-main.30} {{CEREC}: A corpus for entity resolution in email conversations}.
\newblock In \emph{Proceedings of the 28th International Conference on Computational Linguistics}, pages 339--349, Barcelona, Spain (Online). International Committee on Computational Linguistics.

\bibitem[{Eirew et~al.(2021)Eirew, Cattan, and Dagan}]{eirew-etal-2021-wec}
Alon Eirew, Arie Cattan, and Ido Dagan. 2021.
\newblock \href {https://doi.org/10.18653/v1/2021.naacl-main.198} {{WEC}: Deriving a large-scale cross-document event coreference dataset from {W}ikipedia}.
\newblock In \emph{Proceedings of the 2021 Conference of the North American Chapter of the Association for Computational Linguistics: Human Language Technologies}, pages 2498--2510, Online. Association for Computational Linguistics.

\bibitem[{Hamborg et~al.(2019)Hamborg, Zhukova, and Gipp}]{hamborg-2019-automated}
Felix Hamborg, Anastasia Zhukova, and Bela Gipp. 2019.
\newblock \href {https://doi.org/10.1109/JCDL.2019.00036} {Automated identification of media bias by word choice and labeling in news articles}.
\newblock In \emph{2019 ACM/IEEE Joint Conference on Digital Libraries (JCDL)}, pages 196--205.

\bibitem[{Hasler et~al.(2006)Hasler, Orasan, and Naumann}]{hasler-etal-2006-nps}
Laura Hasler, Constantin Orasan, and Karin Naumann. 2006.
\newblock \href {https://aclanthology.org/L06-1325/} {{NP}s for events: Experiments in coreference annotation}.
\newblock In \emph{Proceedings of the Fifth International Conference on Language Resources and Evaluation ({LREC}{'}06)}, Genoa, Italy. European Language Resources Association (ELRA).

\bibitem[{Hong et~al.(2016)Hong, Zhang, O{'}Gorman, Horowit-Hendler, Ji, and Palmer}]{hong-etal-2016-building}
Yu~Hong, Tongtao Zhang, Tim O{'}Gorman, Sharone Horowit-Hendler, Heng Ji, and Martha Palmer. 2016.
\newblock \href {https://doi.org/10.18653/v1/W16-1701} {Building a cross-document event-event relation corpus}.
\newblock In \emph{Proceedings of the 10th Linguistic Annotation Workshop held in conjunction with {ACL} 2016 ({LAW}-X 2016)}, pages 1--6, Berlin, Germany. Association for Computational Linguistics.

\bibitem[{Kazemi et~al.(2025)Kazemi, Fatemi, Bansal, Palowitch, Anastasiou, Mehta, Jain, Aglietti, Jindal, Chen, Dikkala, Tyen, Liu, Shalit, Chiappa, Olszewska, Tay, Tran, Le, and Firat}]{kazemi-etal-2025-big}
Mehran Kazemi, Bahare Fatemi, Hritik Bansal, John Palowitch, Chrysovalantis Anastasiou, Sanket~Vaibhav Mehta, Lalit~K Jain, Virginia Aglietti, Disha Jindal, Peter Chen, Nishanth Dikkala, Gladys Tyen, Xin Liu, Uri Shalit, Silvia Chiappa, Kate Olszewska, Yi~Tay, Vinh~Q. Tran, Quoc~V Le, and Orhan Firat. 2025.
\newblock \href {https://doi.org/10.18653/v1/2025.acl-long.1285} {{BIG}-bench extra hard}.
\newblock In \emph{Proceedings of the 63rd Annual Meeting of the Association for Computational Linguistics (Volume 1: Long Papers)}, pages 26473--26501, Vienna, Austria. Association for Computational Linguistics.

\bibitem[{Minard et~al.(2016)Minard, Speranza, Urizar, Altuna, van Erp, Schoen, and van Son}]{minard-etal-2016-meantime}
Anne-Lyse Minard, Manuela Speranza, Ruben Urizar, Bego{\~n}a Altuna, Marieke van Erp, Anneleen Schoen, and Chantal van Son. 2016.
\newblock \href {https://aclanthology.org/L16-1699/} {{MEANTIME}, the {N}ews{R}eader multilingual event and time corpus}.
\newblock In \emph{Proceedings of the Tenth International Conference on Language Resources and Evaluation ({LREC}'16)}, pages 4417--4422, Portoro{\v{z}}, Slovenia. European Language Resources Association (ELRA).

\bibitem[{Pouran Ben~Veyseh et~al.(2024)Pouran Ben~Veyseh, Lai, Nguyen, Dernoncourt, and Nguyen}]{pouran-ben-veyseh-etal-2024-mcecr}
Amir Pouran Ben~Veyseh, Viet~Dac Lai, Chien Nguyen, Franck Dernoncourt, and Thien Nguyen. 2024.
\newblock \href {https://doi.org/10.18653/v1/2024.findings-naacl.245} {{MCECR}: A novel dataset for multilingual cross-document event coreference resolution}.
\newblock In \emph{Findings of the Association for Computational Linguistics: NAACL 2024}, pages 3869--3880, Mexico City, Mexico. Association for Computational Linguistics.

\bibitem[{Radford(2020)}]{radford-2020-seeing}
Benjamin Radford. 2020.
\newblock \href {https://aclanthology.org/2020.aespen-1.7/} {Seeing the forest and the trees: Detection and cross-document coreference resolution of militarized interstate disputes}.
\newblock In \emph{Proceedings of the Workshop on Automated Extraction of Socio-political Events from News 2020}, pages 35--41, Marseille, France. European Language Resources Association (ELRA).

\bibitem[{Ravenscroft et~al.(2021)Ravenscroft, Clare, Cattan, Dagan, and Liakata}]{ravenscroft-etal-2021-cd}
James Ravenscroft, Amanda Clare, Arie Cattan, Ido Dagan, and Maria Liakata. 2021.
\newblock \href {https://doi.org/10.18653/v1/2021.eacl-main.21} {{CD}{\textasciicircum}2{CR}: Co-reference resolution across documents and domains}.
\newblock In \emph{Proceedings of the 16th Conference of the European Chapter of the Association for Computational Linguistics: Main Volume}, pages 270--280, Online. Association for Computational Linguistics.

\bibitem[{Recasens et~al.(2012)Recasens, Mart{\'i}, and Orasan}]{recasens-etal-2012-annotating}
Marta Recasens, M.~Ant{\`o}nia Mart{\'i}, and Constantin Orasan. 2012.
\newblock \href {https://aclanthology.org/L12-1391/} {Annotating near-identity from coreference disagreements}.
\newblock In \emph{Proceedings of the Eighth International Conference on Language Resources and Evaluation ({LREC}'12)}, pages 165--172, Istanbul, Turkey. European Language Resources Association (ELRA).

\bibitem[{Song et~al.(2018)Song, Bies, Mott, Li, Strassel, and Caruso}]{song-etal-2018-cross}
Zhiyi Song, Ann Bies, Justin Mott, Xuansong Li, Stephanie Strassel, and Christopher Caruso. 2018.
\newblock \href {https://aclanthology.org/L18-1558/} {Cross-document, cross-language event coreference annotation using event hoppers}.
\newblock In \emph{Proceedings of the Eleventh International Conference on Language Resources and Evaluation ({LREC} 2018)}, Miyazaki, Japan. European Language Resources Association (ELRA).

\bibitem[{Tracey et~al.(2022)Tracey, Bies, Getman, Griffitt, and Strassel}]{tracey-etal-2022-study}
Jennifer Tracey, Ann Bies, Jeremy Getman, Kira Griffitt, and Stephanie Strassel. 2022.
\newblock \href {https://aclanthology.org/2022.lrec-1.195/} {A study in contradiction: Data and annotation for {AIDA} focusing on informational conflict in {R}ussia-{U}kraine relations}.
\newblock In \emph{Proceedings of the Thirteenth Language Resources and Evaluation Conference}, pages 1831--1838, Marseille, France. European Language Resources Association.

\bibitem[{Vossen et~al.(2018)Vossen, Ilievski, Postma, and Segers}]{vossen-etal-2018-dont}
Piek Vossen, Filip Ilievski, Marten Postma, and Roxane Segers. 2018.
\newblock \href {https://aclanthology.org/L18-1480/} {Don{'}t annotate, but validate: a data-to-text method for capturing event data}.
\newblock In \emph{Proceedings of the Eleventh International Conference on Language Resources and Evaluation ({LREC} 2018)}, Miyazaki, Japan. European Language Resources Association (ELRA).

\bibitem[{Wang et~al.(2019)Wang, Pruksachatkun, Nangia, Singh, Michael, Hill, Levy, and Bowman}]{Wang2019}
Alex Wang, Yada Pruksachatkun, Nikita Nangia, Amanpreet Singh, Julian Michael, Felix Hill, Omer Levy, and Samuel~R. Bowman. 2019.
\newblock \emph{SuperGLUE: a stickier benchmark for general-purpose language understanding systems}. Curran Associates Inc., Red Hook, NY, USA.

\bibitem[{Wang et~al.(2018)Wang, Singh, Michael, Hill, Levy, and Bowman}]{wang-etal-2018-glue}
Alex Wang, Amanpreet Singh, Julian Michael, Felix Hill, Omer Levy, and Samuel Bowman. 2018.
\newblock \href {https://doi.org/10.18653/v1/W18-5446} {{GLUE}: A multi-task benchmark and analysis platform for natural language understanding}.
\newblock In \emph{Proceedings of the 2018 {EMNLP} Workshop {B}lackbox{NLP}: Analyzing and Interpreting Neural Networks for {NLP}}, pages 353--355, Brussels, Belgium. Association for Computational Linguistics.

\bibitem[{Zhukova et~al.(2026)Zhukova, Hamborg, Donnay, Meuschke, and Gipp}]{Zhukova2026b}
Anastasia Zhukova, Felix Hamborg, Karsten Donnay, Norman Meuschke, and Bela Gipp. 2026.
\newblock \href {https://arxiv.org/pdf/2602.17424} {Diverse word choices, same reference: Annotating lexically-rich cross-document coreference}.
\newblock In \emph{Proceedings of the International Conference on Artificial Intelligence, Computer, Data Sciences and Applications (ACDSA 2026)}, Boracay, Philippines. IEEE.

\end{thebibliography}


\begin{thebibliography}{24}
\expandafter\ifx\csname natexlab\endcsname\relax\def\natexlab#1{#1}\fi

\bibitem[{Ahmed et~al.(2024)Ahmed, Baker, Judge, Reagan, Wright-Bettner, Palmer, and Martin}]{ahmed-etal-2024-linear}
Shafiuddin~Rehan Ahmed, George~Arthur Baker, Evi Judge, Michael Reagan, Kristin Wright-Bettner, Martha Palmer, and James~H. Martin. 2024.
\newblock \href {https://aclanthology.org/2024.lrec-main.920/} {Linear cross-document event coreference resolution with {X}-{AMR}}.
\newblock In \emph{Proceedings of the 2024 Joint International Conference on Computational Linguistics, Language Resources and Evaluation (LREC-COLING 2024)}, pages 10517--10529, Torino, Italia. ELRA and ICCL.

\bibitem[{Ahmed et~al.(2023)Ahmed, Nath, Martin, and Krishnaswamy}]{ahmed-etal-2023-2}
Shafiuddin~Rehan Ahmed, Abhijnan Nath, James~H. Martin, and Nikhil Krishnaswamy. 2023.
\newblock \href {https://doi.org/10.18653/v1/2023.findings-acl.100} {$2*n$ is better than $n^2$: Decomposing event coreference resolution into two tractable problems}.
\newblock In \emph{Findings of the Association for Computational Linguistics: ACL 2023}, pages 1569--1583, Toronto, Canada. Association for Computational Linguistics.

\bibitem[{Bagga and Baldwin(1998)}]{bagga-baldwin-1998-entity}
Amit Bagga and Breck Baldwin. 1998.
\newblock \href {https://doi.org/10.3115/980845.980859} {Entity-based cross-document coreferencing using the vector space model}.
\newblock In \emph{36th Annual Meeting of the Association for Computational Linguistics and 17th International Conference on Computational Linguistics, Volume 1}, pages 79--85, Montreal, Quebec, Canada. Association for Computational Linguistics.

\bibitem[{Barhom et~al.(2019)Barhom, Shwartz, Eirew, Bugert, Reimers, and Dagan}]{barhom-etal-2019-revisiting}
Shany Barhom, Vered Shwartz, Alon Eirew, Michael Bugert, Nils Reimers, and Ido Dagan. 2019.
\newblock \href {https://doi.org/10.18653/v1/P19-1409} {Revisiting joint modeling of cross-document entity and event coreference resolution}.
\newblock In \emph{Proceedings of the 57th Annual Meeting of the Association for Computational Linguistics}, pages 4179--4189, Florence, Italy. Association for Computational Linguistics.

\bibitem[{Bast et~al.(2023)Bast, Hertel, and Walter}]{bast-etal-2023-fast}
Hannah Bast, Matthias Hertel, and Sebastian Walter. 2023.
\newblock \href {https://doi.org/10.18653/v1/2023.acl-demo.37} {Fast whitespace correction with encoder-only transformers}.
\newblock In \emph{Proceedings of the 61st Annual Meeting of the Association for Computational Linguistics (Volume 3: System Demonstrations)}, pages 389--399, Toronto, Canada. Association for Computational Linguistics.

\bibitem[{Caciularu et~al.(2021)Caciularu, Cohan, Beltagy, Peters, Cattan, and Dagan}]{caciularu-etal-2021-cdlm-cross}
Avi Caciularu, Arman Cohan, Iz~Beltagy, Matthew Peters, Arie Cattan, and Ido Dagan. 2021.
\newblock \href {https://doi.org/10.18653/v1/2021.findings-emnlp.225} {{CDLM}: Cross-document language modeling}.
\newblock In \emph{Findings of the Association for Computational Linguistics: EMNLP 2021}, pages 2648--2662, Punta Cana, Dominican Republic. Association for Computational Linguistics.

\bibitem[{Cattan et~al.(2021)Cattan, Eirew, Stanovsky, Joshi, and Dagan}]{cattan-etal-2021-realistic}
Arie Cattan, Alon Eirew, Gabriel Stanovsky, Mandar Joshi, and Ido Dagan. 2021.
\newblock \href {https://doi.org/10.18653/v1/2021.starsem-1.13} {Realistic evaluation principles for cross-document coreference resolution}.
\newblock In \emph{Proceedings of *SEM 2021: The Tenth Joint Conference on Lexical and Computational Semantics}, pages 143--151, Online. Association for Computational Linguistics.

\bibitem[{Chen et~al.(2025)Chen, Li, and Zhu}]{chen-etal-2025-employing}
Xinyu Chen, Peifeng Li, and Qiaoming Zhu. 2025.
\newblock \href {https://doi.org/10.18653/v1/2025.acl-long.1134} {Employing discourse coherence enhancement to improve cross-document event and entity coreference resolution}.
\newblock In \emph{Proceedings of the 63rd Annual Meeting of the Association for Computational Linguistics (Volume 1: Long Papers)}, pages 23272--23286, Vienna, Austria. Association for Computational Linguistics.

\bibitem[{Gao et~al.(2024)Gao, Li, Meng, Li, Zhou, Li, Teng, and Ji}]{gao-etal-2024-enhancing}
Qiang Gao, Bobo Li, Zixiang Meng, Yunlong Li, Jun Zhou, Fei Li, Chong Teng, and Donghong Ji. 2024.
\newblock \href {https://aclanthology.org/2024.lrec-main.523/} {Enhancing cross-document event coreference resolution by discourse structure and semantic information}.
\newblock In \emph{Proceedings of the 2024 Joint International Conference on Computational Linguistics, Language Resources and Evaluation (LREC-COLING 2024)}, pages 5907--5921, Torino, Italia. ELRA and ICCL.

\bibitem[{Held et~al.(2021)Held, Iter, and Jurafsky}]{held-etal-2021-focus}
William Held, Dan Iter, and Dan Jurafsky. 2021.
\newblock \href {https://doi.org/10.18653/v1/2021.emnlp-main.106} {Focus on what matters: Applying discourse coherence theory to cross document coreference}.
\newblock In \emph{Proceedings of the 2021 Conference on Empirical Methods in Natural Language Processing}, pages 1406--1417, Online and Punta Cana, Dominican Republic. Association for Computational Linguistics.

\bibitem[{Hovy et~al.(2013)Hovy, Mitamura, Verdejo, Araki, and Philpot}]{hovy-etal-2013-events}
Eduard Hovy, Teruko Mitamura, Felisa Verdejo, Jun Araki, and Andrew Philpot. 2013.
\newblock \href {https://aclanthology.org/W13-1203/} {Events are not simple: Identity, non-identity, and quasi-identity}.
\newblock In \emph{Workshop on Events: Definition, Detection, Coreference, and Representation}, pages 21--28, Atlanta, Georgia. Association for Computational Linguistics.

\bibitem[{Hsu and Horwood(2022)}]{hsu-horwood-2022-contrastive}
Benjamin Hsu and Graham Horwood. 2022.
\newblock \href {https://doi.org/10.18653/v1/2022.naacl-main.267} {Contrastive representation learning for cross-document coreference resolution of events and entities}.
\newblock In \emph{Proceedings of the 2022 Conference of the North American Chapter of the Association for Computational Linguistics: Human Language Technologies}, pages 3644--3655, Seattle, United States. Association for Computational Linguistics.

\bibitem[{{Linguistic Data Consortium} et~al.(2008)}]{linguistic2008ace}
{Linguistic Data Consortium} et~al. 2008.
\newblock {ACE} (automatic content extraction) english annotation guidelines for entities.
\newblock Technical report, Technical report, Linguistic Data Consortium.

\bibitem[{Luo(2005)}]{luo-2005-coreference}
Xiaoqiang Luo. 2005.
\newblock \href {https://aclanthology.org/H05-1004/} {On coreference resolution performance metrics}.
\newblock In \emph{Proceedings of Human Language Technology Conference and Conference on Empirical Methods in Natural Language Processing}, pages 25--32, Vancouver, British Columbia, Canada. Association for Computational Linguistics.

\bibitem[{McCarthy and Jarvis(2010)}]{McCarthy2010}
Philip~M. McCarthy and Scott Jarvis. 2010.
\newblock \href {https://doi.org/10.3758/BRM.42.2.381} {Mtld, vocd-d, and hd-d: A validation study of sophisticated approaches to lexical diversity assessment}.
\newblock \emph{Behavior Research Methods}, 42(2):381--392.

\bibitem[{Nath et~al.(2024)Nath, Jamil, Ahmed, Baker, Ghosh, Martin, Blanchard, and Krishnaswamy}]{nath-etal-2024-multimodal}
Abhijnan Nath, Huma Jamil, Shafiuddin~Rehan Ahmed, George~Arthur Baker, Rahul Ghosh, James~H. Martin, Nathaniel Blanchard, and Nikhil Krishnaswamy. 2024.
\newblock \href {https://aclanthology.org/2024.lrec-main.1039/} {Multimodal cross-document event coreference resolution using linear semantic transfer and mixed-modality ensembles}.
\newblock In \emph{Proceedings of the 2024 Joint International Conference on Computational Linguistics, Language Resources and Evaluation (LREC-COLING 2024)}, pages 11901--11916, Torino, Italia. ELRA and ICCL.

\bibitem[{Pradhan et~al.(2014)Pradhan, Luo, Recasens, Hovy, Ng, and Strube}]{pradhan-etal-2014-scoring}
Sameer Pradhan, Xiaoqiang Luo, Marta Recasens, Eduard Hovy, Vincent Ng, and Michael Strube. 2014.
\newblock \href {https://doi.org/10.3115/v1/P14-2006} {Scoring coreference partitions of predicted mentions: A reference implementation}.
\newblock In \emph{Proceedings of the 52nd Annual Meeting of the Association for Computational Linguistics (Volume 2: Short Papers)}, pages 30--35, Baltimore, Maryland. Association for Computational Linguistics.

\bibitem[{Recasens et~al.(2010)Recasens, Hovy, and Mart{\'i}}]{recasens-etal-2010-typology}
Marta Recasens, Eduard Hovy, and M.~Ant{\`o}nia Mart{\'i}. 2010.
\newblock \href {https://aclanthology.org/L10-1103/} {A typology of near-identity relations for coreference ({NIDENT})}.
\newblock In \emph{Proceedings of the Seventh International Conference on Language Resources and Evaluation ({LREC}'10)}, Valletta, Malta. European Language Resources Association (ELRA).

\bibitem[{Singh et~al.(2011)Singh, Subramanya, Pereira, and McCallum}]{singh-etal-2011-large}
Sameer Singh, Amarnag Subramanya, Fernando Pereira, and Andrew McCallum. 2011.
\newblock \href {https://aclanthology.org/P11-1080/} {Large-scale cross-document coreference using distributed inference and hierarchical models}.
\newblock In \emph{Proceedings of the 49th Annual Meeting of the Association for Computational Linguistics: Human Language Technologies}, pages 793--803, Portland, Oregon, USA. Association for Computational Linguistics.

\bibitem[{Vilain et~al.(1995)Vilain, Burger, Aberdeen, Connolly, and Hirschman}]{vilain-etal-1995-model}
Marc Vilain, John Burger, John Aberdeen, Dennis Connolly, and Lynette Hirschman. 1995.
\newblock \href {https://aclanthology.org/M95-1005/} {A model-theoretic coreference scoring scheme}.
\newblock In \emph{Sixth Message Understanding Conference ({MUC}-6): Proceedings of a Conference Held in {C}olumbia, {M}aryland, November 6-8, 1995}.

\bibitem[{Yu et~al.(2022)Yu, Yin, and Roth}]{yu-etal-2022-pairwise}
Xiaodong Yu, Wenpeng Yin, and Dan Roth. 2022.
\newblock \href {https://doi.org/10.18653/v1/2022.starsem-1.6} {Pairwise representation learning for event coreference}.
\newblock In \emph{Proceedings of the 11th Joint Conference on Lexical and Computational Semantics}, pages 69--78, Seattle, Washington. Association for Computational Linguistics.

\bibitem[{Zhao et~al.(2023)Zhao, Xue, and Min}]{zhao-etal-2023-cross}
Jin Zhao, Nianwen Xue, and Bonan Min. 2023.
\newblock \href {https://doi.org/10.18653/v1/2023.conll-1.38} {Cross-document event coreference resolution: Instruct humans or instruct {GPT}?}
\newblock In \emph{Proceedings of the 27th Conference on Computational Natural Language Learning (CoNLL)}, pages 561--574, Singapore. Association for Computational Linguistics.

\bibitem[{Zhukova et~al.(2022{\natexlab{a}})Zhukova, Hamborg, Donnay, and Gipp}]{zhukova-2022-xcoref}
Anastasia Zhukova, Felix Hamborg, Karsten Donnay, and Bela Gipp. 2022{\natexlab{a}}.
\newblock \href {https://link.springer.com/chapter/10.1007/978-3-030-96957-8_25} {{XC}oref: Cross-document coreference resolution in the wild}.
\newblock In \emph{Information for a Better World: Shaping the Global Future}, pages 272--291, Cham. Springer International Publishing.

\bibitem[{Zhukova et~al.(2022{\natexlab{b}})Zhukova, Hamborg, and Gipp}]{zhukova-etal-2022-towards}
Anastasia Zhukova, Felix Hamborg, and Bela Gipp. 2022{\natexlab{b}}.
\newblock \href {https://aclanthology.org/2022.lrec-1.522/} {Towards evaluation of cross-document coreference resolution models using datasets with diverse annotation schemes}.
\newblock In \emph{Proceedings of the Thirteenth Language Resources and Evaluation Conference}, pages 4884--4893, Marseille, France. European Language Resources Association.

\end{thebibliography}

\section{Language Resource References}
\label{lr:ref}
\bibliographystylelanguageresource{lrec2026-natbib}
\bibliographylanguageresource{languageresource}

\section*{Appendix}

\subsection{Details on dataset parsing}
\label{sec:dataset_details}

\subsubsection{CD2CR} 
CD2CR (Cross-document Cross-domain Coreference Resolution) addresses the challenge of cross-domain coreference across scientific publications and news articles \citeplanguageresource{ravenscroft-etal-2021-cd}. Unlike previous datasets, which focus on coreference within documents of the same genre and style (e.g., Wikipedia or news), CD2CR incorporates documents from multiple domains, increasing lexical diversity due to differences in language and style. For example, scientific articles employ domain-specific terminology, whereas science-popularization articles simplify language to reach a broader audience. All articles in CD2CR pertain to science and technology topics, resulting in annotated entities that extend beyond common categories, such as person, country/location, and organization, to include specialized entities, such as animals or chemical reactions.

The dataset was originally collected under topics related to science and technology; therefore, we restructured the hierarchy by converting the original topics into subtopics, in accordance with the definitions of topic and subtopic. In the initial version, topics (now treated as subtopics) were not uniquely defined across the train, validation, and test splits. To ensure the correct computation of evaluation metrics and enable unambiguous comparisons across subsets, we assigned unique subtopic identifiers within each split. Furthermore, we removed the former subtopic labels distinguishing news from scientific articles because (i) this distinction did not align with the formal definition of a subtopic, and (ii) each former subtopic contained only a single article (either news or scientific), which is inconsistent with the CDCR task. In the revised version, the dataset comprises a single overarching topic focused on science and technology, with each subtopic representing a specific scientific event and containing two documents: one scientific article and one corresponding news article.

\subsubsection{CEREC}
CEREC (Corpus for Entity Resolution in Email Conversations) annotates entities within email threads \citeplanguageresource{dakle-moldovan-2020-cerec}. In this dataset, each document corresponds to a single email, and each subtopic represents an entire email thread. A key distinguishing feature of CEREC is the semi-structured nature of the documents, which include sender and recipient names, email addresses, subject lines, and message bodies. Coreference relations are annotated across all parts of the emails. 
Additionally, because email is a conversational medium, pronouns vary depending on the author's perspective: for example, the same entity may be referred to as "I" in one email and "you" in another. The dataset includes entity types such as person, organization, location, and digital (i.e., media or references to media stored in digital form). We use the CEREC version employed in the original experiments by \citet{dakle-moldovan-2020-cerec} and refer to it as CEREC\textsubscript{exp}.

Instead of using the full file containing 6,001 annotated mail threads, we rely on the chunked file used for model training in the reference study (train: seed.conll with 43 threads, val: cerec.validation.20.conll with 20 threads, test: cerec.validation.14.conll with 14 threads). We construct each document ID as a combination of a message identifier, the speaker, and a document index that is unique across all data splits, and we exclude all tokens produced by the speaker labeled “SYSTEM.” Coreference chains are defined by the subtopic name, the original coreference identifier within the respective subtopic, and a shortened entity type derived from the extracted and aggregated NER labels of the mentions’ head tokens. The dataset is hierarchically organized into a single overarching topic related to emails, which is subdivided into multiple subtopics corresponding to individual email threads, each containing the associated email documents. 

Despite our effort to correct the text after concatenating the originally tokenized documents, the errors in the text reconstruction prevented the correct concatenation of tokenized emails and resulted in a higher parsing error rate than in the other datasets.

\subsubsection{ECB+}
ECB+ \citeplanguageresource{cybulska-vossen-2014-using} is the most widely used CDCR dataset and serves as a benchmark for ECR. It extends the original EventCorefBank dataset \citeplanguageresource{bejan-harabagiu-2010-unsupervised} by enriching the codebook with entity annotations and adding two new subtopics to each existing topic. The ECB+ codebook primarily focuses on annotating event coreference, while entity coreference is included only when entities function as event attributes, such as participants, locations, or temporal expressions. 
Consequently, events are the main focus of annotation, and entities not associated with events are generally ignored. Furthermore, annotations were typically performed only on the first sentences of documents rather than exhaustively across the entire content.

The news articles in the dataset are hierarchically organized such that each topic corresponds to the original topic\_id, each subtopic is defined as a combination of the topic\_id and its source identifier (ecb or  ECB+), and the documents within each subtopic are represented by enumerated doc\_ids extracted from the original document filenames.

\subsubsection{ECB+METAm}
ECB+METAm \citeplanguageresource{ahmed-etal-2024-generating} addresses the challenges of lexical diversity and ambiguity in event coreference by leveraging GPT-4 to perform constrained metaphoric paraphrasing of annotated mentions in ECB+ validation and test sets. ECB+METAm is generated through a semi-automatic process in which event triggers (i.e., event mentions) from ECB+ are transformed using metaphoric paraphrasing. For example, the trigger word "killing" from ECB+ is paraphrased as "snuffing out the flame of life." We use ECB+METAm instead of ECB+META\textsubscript{1} because this version exhibits higher lexical diversity. During re-parsing ECB+METAm, several inconsistencies occurred due to the paraphrase errors on a sentence level: paraphrasing of event occurrences also triggered paraphrases of their entity attributes, but the annotated entity mention remained unchanged and couldn’t be mapped to the modified sentence. In this case, an entity mention was skipped as not found in the paraphrased sentence. 

\subsubsection{FCC-T}
FCC-T is a tokenized version of the Football Coreference Corpus (FCC) dataset, designed for studying cross-subtopic event coreference relations \citeplanguageresource{bugert-2020-breaking, bugert-2021-generalizing}. In FCC-T, each event corresponds to a football game within a set of pre-selected tournaments. Similar to ECB+, FCC-T annotates the semantic roles associated with each event, such as participants, locations, and time, yet it does not identify coreference clusters among these roles. We parse but do not output these mentions (optionally possible).

The dataset articles are hierarchically organized into a single overarching topic focused on football matches, which is subdivided into multiple subtopics representing distinct seminal football events; each subtopic contains the corresponding news articles describing that event. 

\subsubsection{GVC}
GVC (Gun Violence Corpus) is an ECR dataset that focuses on coreference clusters within the gun violence domain \citeplanguageresource{vossen-etal-2018-dont}. The dataset annotates mentions and clusters them according to predefined event classes, such as "injury" or "missing", which characterize the domain. Each event is specified by its location, time, and information about the victims. Actors involved in violent actions are included in the event if their reference terms share a root with the action (e.g., "killer" – "kill"). Because the codebook restricts the types of events that can be annotated, the resulting clusters exhibit high lexical ambiguity and substantial overlap across mentions.

The assignment of documents to subtopics, as well as the allocation of subtopics to the train, validation, and test splits, follows the structure provided in the GitHub repository of \citet{bugert-2021-generalizing}. Mentions with a cluster\_ID of 0 are treated as singletons and are therefore assigned unique cluster identifiers. The dataset is hierarchically organized into a single overarching topic on gun violence, which is subdivided into multiple subtopics representing specific gun violence–related events; each subtopic contains the corresponding news articles reporting on these incidents.

\subsubsection{HyperCoref}
HyperCoref is the largest published event-level CDCR dataset, collected from hyperlinks in news articles \citeplanguageresource{bugert-gurevych-2021-event}. By leveraging hyperlinks, HyperCoref accelerates the otherwise time-consuming annotation process. The dataset also aims to expand the number of topics, which are typically limited in other corpora, and to overcome the constraints imposed by the Wikipedia community regarding which events are deemed noteworthy for reporting in Wikinews. As a result, HyperCoref provides a more diverse set of documents across both topics and lexicons. 

Although the original HyperCoref paper states that the dataset contains only event annotations, our analysis indicates that it also contains several entity mentions. Accordingly, we construct a file containing entity mentions whose head tokens have part-of-speech tags in the set {ADJ, ADV, ADP, NUM, NOUN, PRON, PROPN}, thereby systematically capturing entity-level annotations identified in the data. Since the coreference relations were built on the hyperlinks, to indicate the loose nature of these links, we indicate in \Cref{tab:annot} the near-identity relations.

We adopt the original train/val/test split and preprocess the data to closely replicate the experimental setup described in the original paper. Specifically, we exclude subtopics with a singleton only, downsample the training sets to 25K mentions per outlet, and reduce the validation sets to 1,700 mentions for ABC and 2,400 for BBC. During downsampling, we prioritize retaining larger clusters before smaller ones and randomly select clusters only when necessary to meet the predefined mention limits, thereby minimizing the number of singleton clusters. In the test sets, we remove subtopics containing only one mention to ensure valid subtopic-level evaluation later. We preserve the original maximum-span annotation style of HyperCoref and maintain the initial document tokenization without re-parsing the documents; instead, we align the re-parsed mentions with the original tokenized texts.

To adapt HyperCoref to the uCDCR topic–subtopic–document hierarchy, we derive topic and subtopic labels from the original document URLs. The topic corresponds to the highest-level folder name in the URL, with its first letter capitalized to ensure consistency across outlets. Subtopics are constructed by combining the outlet name with the second URL component: for ABC News, we use the second component when it is capitalized; for BBC, we select the first non-numeric component and retain the segment preceding any dash. If URL parsing fails to get the second component, the subtopic defaults to the topic name. The resulting structure organizes the dataset hierarchically into topics (highest-level URL folders), subtopics (outlet name plus subsequent folder component), and documents (individual news article URLs).

\subsubsection{MEANTIME}
MEANTIME explores a multilingual setup for both event- and entity-level CDCR \citeplanguageresource{minard-etal-2016-meantime}. The dataset consists of Wikinews articles on four economic and financial topics, each translated into Dutch, Spanish, and Italian. Entities are annotated independently from events, following ACE guidelines \cite{linguistic2008ace}. 
Similar to ECB+, events are annotated only when the four discourse elements are identical. Unlike ECB+, MEANTIME provides thorough annotation across the entire document, identifying all markables. The annotation process began with within-document entity and event chains, which were subsequently merged into cross-document clusters. 

To align MEANTIME with the three-level CDCR structure described above, each article is treated as a subtopic, with its four language versions representing the documents of that subtopic, thus enabling MEANTIME to include both within- and cross-subtopic annotations. Although the MEANTIME\textsubscript{eng} version is limited to within-document coreference resolution due to the absence of the other three language versions per subtopic, the number of mentions per document is four times greater than in ECB+ (see \Cref{tab:general_stats}), which is sufficient for the CDCR task.

We propose a train–validation–test split for the MEANTIME dataset, with partitioning at the topic level: two topics to the training set and one topic each to the validation and test sets. The mapping between topic identifiers and topic names is as follows: 0 – corpus\_airbus, 1 – corpus\_apple, 2 – corpus\_gm, and 3 – corpus\_stock.

\subsubsection{NewsWCL50r}
NewsWCL50r (News Word Choice and Labeling 50 reannotated) \citeplanguageresource{Zhukova2026b} annotates both event- and entity-level CDCR by identifying coreference-like relations between phrases exhibiting bias through word choice or labeling or entities or events that may be framed\citeplanguageresource{hamborg-2019-automated}. Coreference relations between bias-loaded phrases and their corresponding entities or events vary in the degree of identity, ranging from looser relations (e.g., "DACA recipients" – "kids that have done nothing wrong") to context-dependent synonyms and regular strict identity. Events and entities were annotated independently using a maximum-span approach, where markables were extended as needed to capture the full sentiment of the mentions. 

The dataset comprises 10 topics without an explicit subtopic level. However, according to our formal definition, subtopics group event-related articles, whereas topics aggregate multiple subtopics. To align the dataset with the required topic/subtopic/document hierarchy, we convert each original topic into a subtopic and assign it to a newly introduced topic with the same identifier. Additionally, because the original texts use newline delimiters to separate paragraphs, we preserve these delimiters as tokens in the dataset. To prevent parsing issues in the CoNLL format, newline characters are stored as \textbackslash\textbackslash n; consequently, reconstruction of the original plain text requires replacing \textbackslash\textbackslash n with \textbackslash n.

\subsubsection{NP4E}
NP4E (Noun Phrases for Events) annotates entities that serve as participants, locations, or temporal expressions, i.e., potential event arguments, as well as noun phrases representing actions \citeplanguageresource{hasler-etal-2006-nps}. The dataset comprises four subtopics within the broader domain of terrorism and security. Coreference relations in NP4E are defined to include identity, synonymy, and lexical generalization or specialization. The annotation is highly detailed and includes numerous nested mentions. To establish the CDCR setup, we applied semi-automated heuristics to cluster mentions from the annotated coreference chains, and we refer to this version as NP4E\textsubscript{cd}.

Because event annotation in the dataset was restricted to five predefined event types and event coreference clusters were not provided for all topics, we extracted only entity coreference clusters. In contrast to most CDCR datasets, event-denoting noun phrases (e.g., an attack) are annotated as entities rather than as events. Furthermore, since the MMAX format does not include an explicit tag to link coreference chains across event-related documents into cross-document clusters, we reconstruct CDCR clusters using a simple yet robust heuristic: chains across documents are merged if at least two non-pronominal mentions, or their head tokens, are identical. Additionally, we allow merging in cases where there is a single overlap involving a proper noun mention.

The news articles in the dataset are hierarchically organized into a single overarching topic encompassing incidents of bombings, explosions, and kidnappings, which is further divided into subtopics representing individual events; each subtopic contains the corresponding news articles describing that event. We also propose a train/val/test split for NP4E by partitioning at the subtopic level, assigning three subtopics to the training set and one subtopic each to the validation and test sets. The mapping between subtopic identifiers and topic names is as follows: 0 – bukavu, 1 – china, 2 – israel, 3 – peru, and 4 – tajikistan.

\subsubsection{NIdent}
NIdent is an entity-only CDCR dataset with annotated identity and near-identity relations \citeplanguageresource{recasens-etal-2010-typology, recasens-etal-2012-annotating}. The dataset captures varying strengths of reference relations, including references that express opinion, are context-dependent, or involve metonymy, meronymy, or set relations. NIdent reannotates the NP4E text collection, including entities such as participants, locations, and actions expressed as nouns. 

Because the original MMAX format lacks an explicit tag to link within-document coreference chains across event-related documents into cross-document clusters, we conducted manual annotation to merge document-level chains into cross-document coreference clusters. The annotation was done when each mention was provided with both the name of the coreference chain and the context of each mention to decide if two chains were to be merged or not. We refer to this CDCR version as NIdent\textsubscript{en-cd}. The topic organization and dataset split are identical to NP4E. 

\subsubsection{WEC-Eng}
WEC-Eng (Wikipedia Event Corpus for English) is a large-scale dataset of events extracted from Wikipedia articles \citeplanguageresource{eirew-etal-2021-wec}. In this dataset, an event is defined as a Wikipedia page referenced by other pages through anchor texts and hyperlinks. Subtopics, or pivot events, correspond to Wikipedia pages whose infoboxes belong to predefined event-related categories, such as Awards or Meetings. Documents in the dataset are the pages that mention these pivot events. In WEC-Eng, pivot events serve a dual role: they act as subtopics and as the names of coreference clusters. 

We adopt the topic names from Table 7 of the WEC paper and reconstruct the assignment of subtopics, defined as the Wikipedia titles of the coreference chains representing seminal (seed) events, to these topics. To this end, we develop a heuristic that extracts the Wikipedia categories associated with each subtopic page and matches them to the predefined topics based on category overlap. The resulting assignments are subsequently manually validated to ensure consistency and correctness. The finalized mapping of subtopics to topics is provided in the WEC-Eng/topics directory. 
The final hierarchical organization of the dataset follows a topic–subtopic–document structure, where each topic contains subtopics corresponding to coreference chain names (i.e., so-called seminal events), and each subtopic comprises the associated Wikipedia articles.

Because a single document may contain mentions belonging to multiple coreference chains (i.e., multiple seminal events), it is necessary to enforce a unique subtopic assignment per document. We therefore assign each document to the first subtopic listed within the corresponding topic file in the WEC-Eng/topics folder. This design allows for cross-subtopic coreference links and is consistent with the experimental setups used in MEANTIME and FCC. To create CoNLL files, we reconstruct Wikipedia articles by merging paragraphs derived from the contextual spans of annotated mentions. As sentence boundaries are not provided in the original data, we assign one sentence ID per paragraph (i.e., per mention context). To align with the preprocessing logic applied to other datasets, each mention context is further restricted to a window of ±100 tokens around the mention’s first token.

\subsection{Additional experiment results}
\label{sec:add_results}

\paragraph{General statistics}

\Cref{tab:parsing_rate} provides information on the parsing rate of the uCDCR datasets relative to the originally released mentions in each dataset. \Cref{tab:pos}, \Cref{tab:ner}, and \Cref{tab:type} report about the POS tag, NER, and mention type distribution across the datasets. \Cref{tab:full_stats} provides a detailed breakdown of the dataset composition for each dataset split, and \Cref{tab:topics1} and \Cref{tab:topics1} list all topics in uCDCR.

\begin{table}[h]
\centering
\scriptsize
\begin{tabular}{|l|r|r|r|}
\hline
\textbf{Dataset} & \textbf{Original} & \textbf{Parsed} & \textbf{Rate} \\
\hline
CD2CR & 7602 & 7597 & 99.93 \\
CEREC\textsubscript{exp} & 7219  & 7080 & 98.07 \\	
ECB+ & 15122 & 15051 & 99.53 \\
ECB+METAm & 6556 & 6348 & 96.83 \\
FCC-T & 3563 & 3561 & 99.94 \\
GVC & 7298 & 7284 & 99.81 \\
HyperCoref\textsubscript{exp} & 60687 & 60401 & 99.53  \\		
MEANTIME\textsubscript{eng} & 6545 & 6506 & 99.40 \\	
NewsWCL50r & 6531 & 6531 & 100.00 \\
NIdent\textsubscript{en-cd} & 13095  & 12988 &  99.18 \\		
NP4E\textsubscript{cd} & 6574 & 6559 & 99.77 \\		
WEC-Eng & 43672 & 43672 & 100.00 \\
\hline
\end{tabular}
\caption{Mentions original and re-parsed. }
\label{tab:parsing_rate}
\end{table}

\begin{table*}[h]
\centering
{\fontsize{5.5}{7}\selectfont
\begin{tabular}{|l|llllllllllll|}
\hline
\textbf{POS} & \textbf{CD2CR} & \textbf{CEREC} & \textbf{ECB+} & \textbf{ECB+METAm} & \textbf{FCC-T} & \textbf{GVC} & \textbf{HyperCoref} & \textbf{MEANTIME} & \textbf{NewsWCL50r} & \textbf{NIdent} & \textbf{NP4E} & \textbf{WEC-Eng} \\
\hline
ADJ & 140 & 44 & 271 & 90 & 338 & 528 & 1245 & 115 & 92 & 116 & 25 & 265 \\
ADP & 52 & 43 & 645 & 508 & 72 & 8 & 2506 & 34 & 65 & 254 & 37 & 207 \\
ADV & 22 & 11 & 28 & 11 & 5 & 106 & 392 & 7 & 11 & 20 & 0 & 45 \\
AUX & 2 & 3 & 12 & 7 & 5 & 3 & 835 & 121 & 14 & 4 & 1 & 87 \\
CCONJ & 1 & 0 & 1 & 2 & 0 & 0 & 14 & 0 & 0 & 0 & 0 & 0 \\
DET & 19 & 6 & 18 & 5 & 2 & 7 & 57 & 17 & 1 & 42 & 34 & 24 \\
INTJ & 0 & 6 & 0 & 0 & 0 & 0 & 1 & 0 & 0 & 0 & 0 & 1 \\
NOUN & 5968 & 609 & 5735 & 2505 & 1432 & 3373 & 21423 & 2148 & 2897 & 7273 & 2647 & 19667 \\
NUM & 154 & 24 & 254 & 105 & 96 & 0 & 1276 & 69 & 14 & 178 & 43 & 70 \\
PART & 0 & 1 & 1 & 0 & 0 & 0 & 9 & 0 & 0 & 1 & 0 & 0 \\
PRON & 200 & 2410 & 635 & 252 & 5 & 268 & 101 & 601 & 8 & 1395 & 1222 & 20 \\
PROPN & 855 & 3551 & 3366 & 1397 & 1096 & 148 & 7813 & 1711 & 2676 & 3493 & 2487 & 20726 \\
PUNCT & 19 & 6 & 11 & 1 & 0 & 0 & 48 & 2 & 0 & 30 & 0 & 6 \\
SCONJ & 1 & 3 & 23 & 11 & 0 & 4 & 14 & 9 & 3 & 5 & 1 & 2 \\
SPACE & 0 & 0 & 0 & 0 & 0 & 0 & 7 & 0 & 0 & 0 & 0 & 0 \\
SYM & 1 & 0 & 0 & 0 & 0 & 0 & 38 & 0 & 0 & 0 & 0 & 0 \\
VERB & 156 & 24 & 4048 & 1452 & 508 & 2837 & 24622 & 1672 & 743 & 177 & 62 & 2507 \\
X & 7 & 339 & 3 & 2 & 2 & 2 & 0 & 0 & 7 & 0 & 0 & 45 \\
\hline
\end{tabular}
}
\caption{POS tag distribution of the mentions' head tokens across the datasets.}
\label{tab:pos}
\end{table*}

\begin{table*}[h]
\centering
\tiny
\begin{tabular}{|l|llllllllllll|}
\hline
\textbf{NER} & \textbf{CD2CR} & \textbf{CEREC} & \textbf{ECB+} & \textbf{ECB+METAm} & \textbf{FCC-T} & \textbf{GVC} & \textbf{HyperCoref} & \textbf{MEANTIME} & \textbf{NewsWCL50r} & \textbf{NIdent} & \textbf{NP4E} & \textbf{WEC-Eng} \\
\hline
CARDINAL   & 129 & 5 & 129 & 54 & 10 & 0 & 764 & 50 & 26 & 136 & 30 & 100 \\
DATE & 162 & 6 & 607 & 254 & 100 & 16 & 888 & 3 & 12 & 652 & 281 & 13 \\
EVENT & 2 & 4 & 28 & 0 & 682 & 1 & 244 & 1 & 0 & 6 & 2 & 1577 \\
FAC & 7 & 11 & 35 & 25 & 9 & 2 & 167 & 9 & 16 & 18 & 3 & 120 \\
GPE & 185 & 269 & 279 & 145 & 22 & 6 & 905 & 228 & 785 & 1243 & 872 & 7 \\
LANGUAGE & 2 & 0 & 0 & 0 & 0 & 0 & 0 & 0 & 0 & 6 & 2 & 0 \\
LAW & 2 & 1 & 8 & 0 & 68 & 0 & 55 & 1 & 1 & 0 & 0 & 125 \\
LOC & 47 & 15 & 75 & 56 & 11 & 0 & 83 & 11 & 52 & 44 & 17 & 1 \\
MONEY & 8 & 0 & 50 & 28 & 4 & 0 & 156 & 12 & 1 & 2 & 1 & 0 \\
NORP & 19 & 34 & 36 & 17 & 4 & 1 & 205 & 56 & 41 & 245 & 143 & 4 \\
O & 6588 & 4108 & 11300 & 4769 & 2520 & 7208 & 51278 & 4835 & 3874 & 9263 & 4163 & 37484 \\
ORDINAL & 3 & 2 & 25 & 10 & 2 & 0 & 176 & 2 & 2 & 8 & 1 & 61 \\
ORG & 245 & 765 & 1035 & 494 & 84 & 23 & 1493 & 1073 & 463 & 614 & 412 & 3400 \\
PERCENT & 94 & 0 & 5 & 2 & 0 & 0 & 17 & 5 & 2 & 13 & 1 & 0 \\
PERSON & 84 & 1832 & 1125 & 390 & 37 & 19 & 3575 & 158 & 1237 & 618 & 568 & 35 \\
PRODUCT & 8 & 12 & 24 & 17 & 5 & 3 & 54 & 55 & 3 & 31 & 21 & 574 \\
QUANTITY & 5 & 0 & 44 & 13 & 0 & 0 & 24 & 2 & 4 & 20 & 16 & 3 \\
TIME & 3 & 0 & 194 & 67 & 0 & 0 & 90 & 2 & 0 & 53 & 10 & 26 \\
\makecell[l]{WORK\_OF\_ART} & 4 & 16 & 52 & 7 & 3 & 5 & 227 & 3 & 12 & 16 & 16 & 142 \\
\hline
\end{tabular}
\caption{NER distribution of the mentions' head tokens across the datasets.}
\label{tab:ner}
\end{table*}

\begin{table*}[h]
\centering
\tiny
\begin{tabular}{|l|llllllllllll|}
\hline
\textbf{Type} & \textbf{CD2CR} & \textbf{CEREC} & \textbf{ECB+} & \textbf{ECB+METAm} & \textbf{FCC-T} & \textbf{GVC} & \textbf{HyperCoref} & \textbf{MEANTIME} & \textbf{NewsWCL50r} & \textbf{NIdent} & \textbf{NP4E} & \textbf{WEC-Eng} \\
\hline
ACTION & 0 & 0 & 6707 & 2942 & 0 & 0 & 0 & 0 & 461 & 0 & 0 & 0 \\
ACTOR & 0 & 0 & 0 & 0 & 0 & 0 & 0 & 0 & 1886 & 0 & 0 & 0 \\
CARDINAL & 125 & 17 & 0 & 0 & 0 & 0 & 115 & 0 & 0 & 1171 & 189 & 0 \\
COUNTRY & 0 & 0 & 0 & 0 & 0 & 0 & 0 & 0 & 1338 & 0 & 0 & 0 \\
DATE & 152 & 7 & 0 & 0 & 0 & 0 & 158 & 0 & 0 & 839 & 295 & 0 \\
EVENT & 2 & 11 & 0 & 0 & 3561 & 7284 & 53101 & 1870 & 647 & 20 & 2 & 43672 \\
FAC & 5 & 48 & 0 & 0 & 0 & 0 & 28 & 0 & 0 & 17 & 0 & 0 \\
FINANCE & 0 & 0 & 0 & 0 & 0 & 0 & 0 & 490 & 0 & 0 & 0 & 0 \\
GPE & 186 & 428 & 0 & 0 & 0 & 0 & 111 & 0 & 0 & 2034 & 1238 & 0 \\
GRAMMATICAL & 0 & 0 & 0 & 0 & 0 & 0 & 0 & 301 & 0 & 0 & 0 & 0 \\
GROUP & 0 & 0 & 0 & 0 & 0 & 0 & 0 & 0 & 546 & 0 & 0 & 0 \\
HUMAN & 0 & 0 & 4571 & 2007 & 0 & 0 & 0 & 0 & 0 & 0 & 0 & 0 \\
LANGUAGE & 2 & 0 & 0 & 0 & 0 & 0 & 0 & 0 & 0 & 9 & 3 & 0 \\
LAW & 2 & 15 & 0 & 0 & 0 & 0 & 4 & 0 & 0 & 0 & 0 & 0 \\
LOC & 44 & 14 & 1169 & 460 & 0 & 0 & 20 & 388 & 0 & 154 & 116 & 0 \\
MISC & 0 & 0 & 0 & 0 & 0 & 0 & 0 & 0 & 665 & 0 & 0 & 0 \\
MIXTURE & 0 & 0 & 0 & 0 & 0 & 0 & 0 & 3 & 0 & 0 & 0 & 0 \\
MONEY & 7 & 0 & 0 & 0 & 0 & 0 & 7 & 0 & 0 & 4 & 2 & 0 \\
NEG & 0 & 0 & 116 & 76 & 0 & 0 & 0 & 0 & 0 & 0 & 0 & 0 \\
NON & 0 & 0 & 1401 & 413 & 0 & 0 & 0 & 0 & 0 & 0 & 0 & 0 \\
NORP & 20 & 105 & 0 & 0 & 0 & 0 & 29 & 0 & 0 & 619 & 323 & 0 \\
OBJECT & 0 & 0 & 0 & 0 & 0 & 0 & 0 & 0 & 768 & 0 & 0 & 0 \\
ORDINAL & 3 & 3 & 0 & 0 & 0 & 0 & 20 & 0 & 0 & 6 & 0 & 0 \\
ORG & 202 & 1751 & 0 & 0 & 0 & 0 & 263 & 1444 & 220 & 1586 & 1087 & 0 \\
OTHER & 6677 & 959 & 0 & 0 & 0 & 0 & 5612 & 0 & 0 & 4580 & 1825 & 0 \\
PERCENT & 91 & 0 & 0 & 0 & 0 & 0 & 0 & 0 & 0 & 12 & 2 & 0 \\
PERSON & 67 & 3595 & 0 & 0 & 0 & 0 & 839 & 559 & 0 & 1682 & 1327 & 0 \\
PRODUCT & 5 & 63 & 0 & 0 & 0 & 0 & 3 & 973 & 0 & 145 & 94 & 0 \\
QUANTITY & 5 & 0 & 0 & 0 & 0 & 0 & 3 & 0 & 0 & 29 & 23 & 0 \\
SPEECH & 0 & 0 & 0 & 0 & 0 & 0 & 0 & 478 & 0 & 0 & 0 & 0 \\
TIME & 1 & 0 & 1087 & 450 & 0 & 0 & 14 & 0 & 0 & 33 & 2 & 0 \\
WORK & 1 & 64 & 0 & 0 & 0 & 0 & 74 & 0 & 0 & 48 & 31 & 0 \\
\hline
\end{tabular}
\caption{Mention type distribution across the datasets (based on the mentions' head tokens). FAC - Facility, GPE - Geo-Political Entity, LOC - Location, MISC - Miscellaneous, NEG - Negative action, 
NON - Non-human participant, NORP - Nationalities or religious or Political groups, ORG - Organization, SPEECH - Speech cognitive, WORK - Work of art }
\label{tab:type}
\end{table*}

\begin{figure*}[h]
    \centering
    \includegraphics[width=0.9\linewidth]{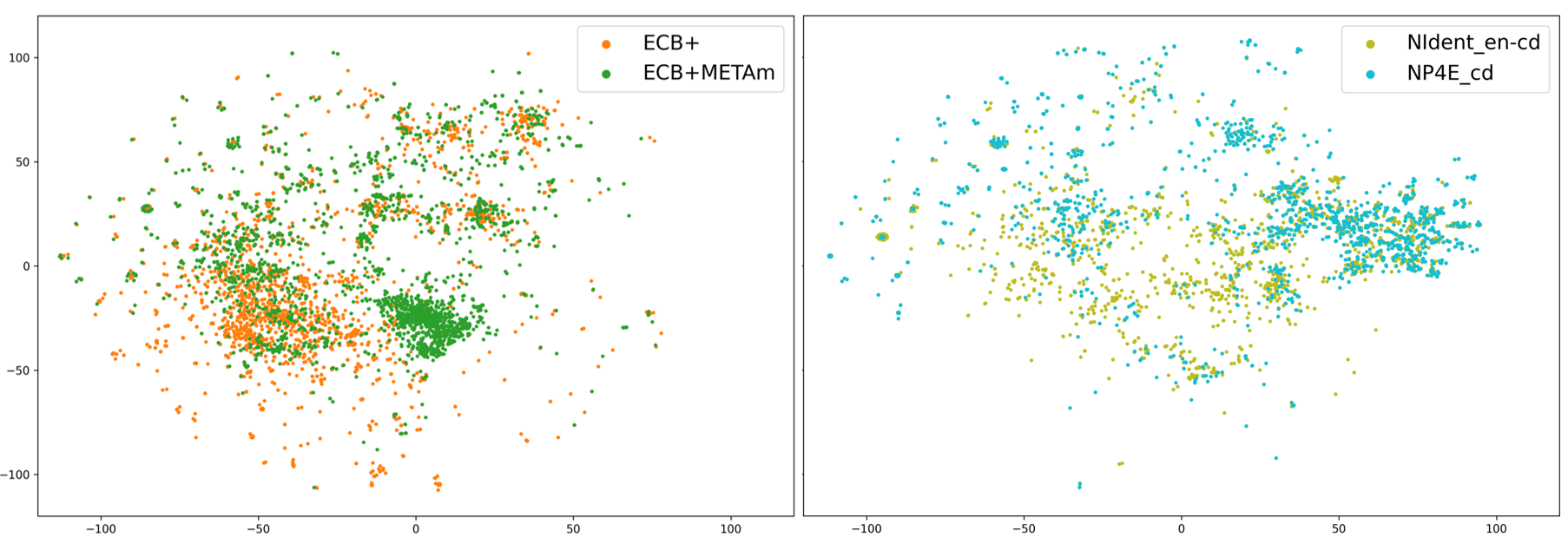}
    \caption{A difference in the t-SNE representation across the datasets with the same source documents but different annotation guidelines: when the same set of texts contained paraphrased mentions (ECB+ vs. ECB+METAm) or were reannotated (NIdent vs. NP4E).}
    \label{fig:reannot_comparision}
\end{figure*}

\begin{figure*}[h]
    \centering
    \includegraphics[width=0.9\linewidth]{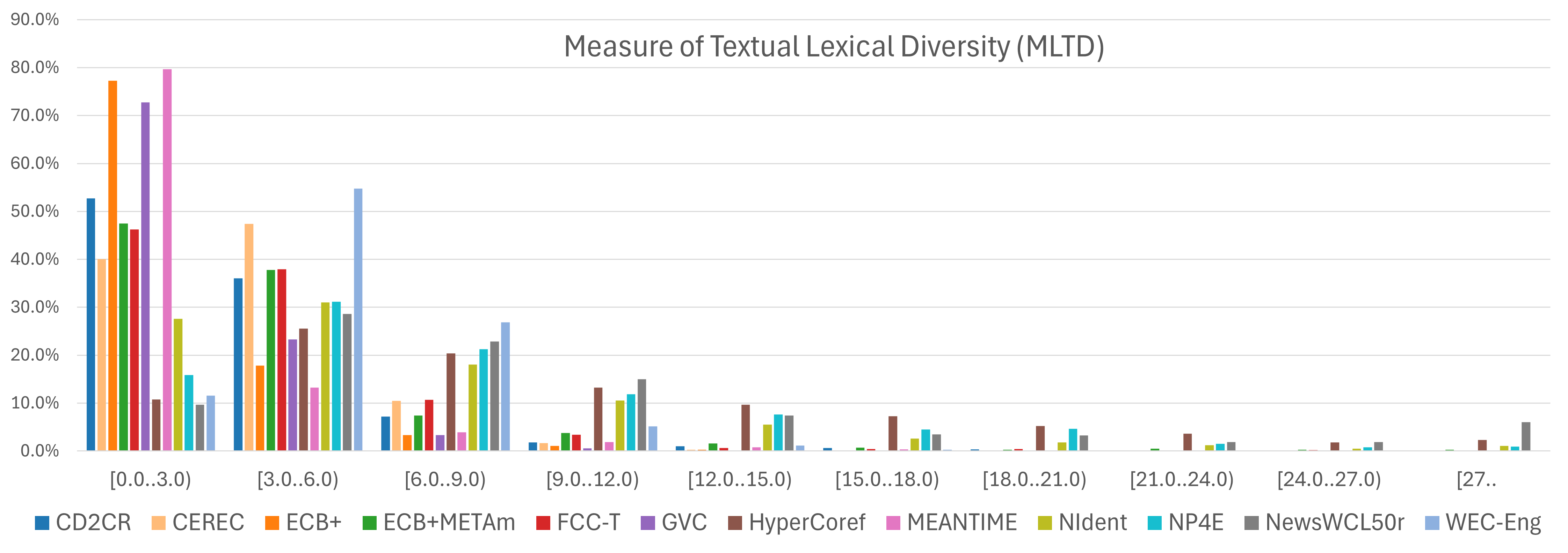}
    \caption{A distribution of the MLTD for the lexical diversity across all datasets. The majority of the datasets contain coreference chains with values between 0 and 9. Only HyperCoref\textsubscript{exp}, NewsWCL50r, NP4E\textsubscript{cd}, and NIdent\textsubscript{en-cd} have some coreference chains with values higher than 20.}
    \label{fig:histogram_MLTD}
\end{figure*}

\begin{figure*}[h]
    \centering
    \includegraphics[width=0.9\linewidth]{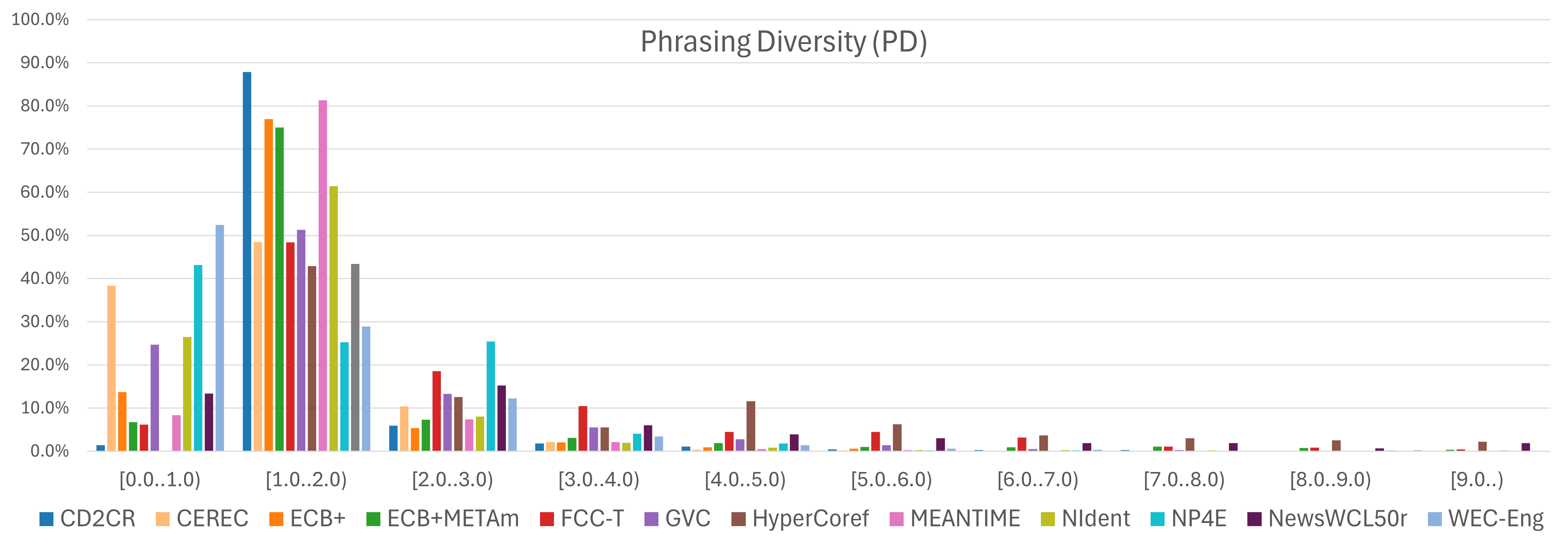}
    \caption{A distribution of the PD metric for the lexical diversity across all datasets. The majority of the datasets contain coreference chains with values between 1 and 2. Only HyperCoref\textsubscript{exp} and NewsWCL50r have some coreference chains with values higher than 9.}
    \label{fig:histogram_PD}
\end{figure*}

\paragraph{Lexical diversity and ambiguity}
\Cref{fig:reannot_comparision} shows the difference in the mention distribution when the same set of texts was reannotated (NIdent vs. NP4E) or contained paraphrased mentions (ECB+ vs. ECB+METAm). \Cref{fig:histogram_PD} and \Cref{fig:histogram_MLTD} show the difference in the distribution of lexical diversity measured by PD and MLTD across the datasets. 

\begin{table*}[h]
\scriptsize
\centering
\begin{tabular}{|l|l|r|r|r|r|r|r|r|r|}
\hline
\textbf{Dataset} & \textbf{Split} & \textbf{Topics} & \textbf{Subtopics} & \textbf{Docs} & \textbf{Tokens} & \textbf{Coref. chain, all} & \textbf{Coref. chain} & \textbf{Singletons} & \textbf{Mentions} \\
\hline
\multirow{3}{*}{CD2CR} & test & 1 & 43 & 86 & 13707 & 756 & 101 & 655 & 1175 \\ 
& train & 1 & 150 & 300 & 49764 & 3221 & 426 & 2795 & 4602 \\ 
& val & 1 & 71 & 142 & 23130 & 1245 & 199 & 1046 & 1820 \\
\hline
\multirow{3}{*}{CEREC\textsubscript{exp}} & test & 1 & 14 & 87 & 11428 & 303 & 154 & 149 & 1267 \\ 
& train & 1 & 43 & 234 & 33371 & 750 & 535 & 215 & 3719 \\
& val & 1 & 20 & 135 & 17542 & 422 & 300 & 122 & 2094 \\ 
\hline
\multirow{3}{*}{ECB+} & test & 10 & 20 & 206 & 86152 & 1408 & 379 & 1029 & 3820 \\ & train & 25 & 50 & 574 & 432856 & 2810 & 877 & 1933 & 8521 \\ & val & 8 & 16 & 196 & 108821 & 737 & 254 & 483 & 2710 \\ \hline
\multirow{2}{*}{ECB+METAm} & test & 10 & 20 & 206 & 89641 & 1376 & 367 & 1009 & 3729 \\ & val & 8 & 16 & 196 & 94789 & 719 & 247 & 472 & 2619 \\ 
\hline
\multirow{3}{*}{FCC-T} & test & 1 & 41 & 122 & 104275 & 191 & 104 & 87 & 1208 \\
& train & 1 & 107 & 202 & 166609 & 259 & 151 & 108 & 1603 \\
& val & 1 & 35 & 104 & 84068 & 136 & 77 & 59 & 750 \\
\hline
\multirow{3}{*}{GVC} & test & 1 & 34 & 74 & 23872 & 250 & 150 & 100 & 1008 \\
& train & 1 & 170 & 358 & 137920 & 1158 & 737 & 421 & 5299 \\
& val & 1 & 37 & 78 & 23863 & 271 & 157 & 114 & 977 \\
\hline
\multirow{3}{*}{HyperCoref\textsubscript{exp}} & test & 20 & 77 & 3381 & 1766034 & 3721 & 671 & 3050 & 4759 \\
& train & 34 & 309 & 33052 & 25743052 & 5477 & 5309 & 168 & 49777 \\
& val & 23 & 101 & 4505 & 2254650 & 3904 & 1253 & 2651 & 5865 \\
\hline
\multirow{3}{*}{MEANTIME\textsubscript{eng}} & test & 1 & 30 & 30 & 14142 & 686 & 136 & 550 & 1256 \\
& train & 2 & 60 & 60 & 27237 & 1522 & 374 & 1148 & 3355 \\
& val & 1 & 30 & 30 & 11625 & 756 & 184 & 572 & 1895 \\
\hline
\multirow{2}{*}{NewsWCL50r} & test & 6 & 6 & 30 & 29721 & 288 & 205 & 83 & 3791 \\ & val & 4 & 4 & 20 & 19871 & 145 & 126 & 19 & 2740 \\
\hline
\multirow{3}{*}{NIdent\textsubscript{en-cd}}& test & 1 & 1 & 20 & 10440 & 347 & 183 & 164 & 2630 \\
& train & 1 & 3 & 55 & 27678 & 1316 & 698 & 618 & 7178 \\ 
& val & 1 & 1 & 18 & 12240 & 800 & 307 & 493 & 3180 \\ 
\hline
\multirow{3}{*}{NP4E\textsubscript{cd}} & test & 1 & 1 & 20 & 10421 & 136 & 136 & 0 & 1440 \\
& train & 1 & 3 & 55 & 27746 & 332 & 332 & 0 & 3375 \\ 
& val & 1 & 1 & 19 & 13062 & 199 & 199 & 0 & 1744 \\ \hline
\multirow{3}{*}{WEC-Eng} & test & 15 & 322 & 1803 & 221735 & 322 & 306 & 16 & 1893 \\
& train & 18 & 6815 & 34132 & 4685903 & 7042 & 6043 & 999 & 40529 \\
& val & 16 & 233 & 1194 & 146851 & 233 & 216 & 17 & 1250 \\
\hline
\end{tabular}
\caption{Overview of the dataset splits of all datasets.}
\label{tab:full_stats}
\end{table*}

\paragraph{Baseline performance analysis}
\Cref{tab:full_performance}  reports all coreference metrics on a subtopic and topic level and on the mention types.


\begin{table*}[h]
\scriptsize
\centering
\begin{tabular}{|l|l|l|lll|lll|lll|c|}
\hline
\multirow{2}{*}{Dataset} &  &  & \multicolumn{3}{c|}{MUC} & \multicolumn{3}{c|}{B3} & \multicolumn{3}{c|}{CEAF\_e} & \multirow{2}{*}{F1 CoNLL} \\
\cline{4-12}
 & Level & type & R & P & F1 & R & P & F1 & R & P & F1 &  \\
 \hline
CD2CR & \multirow{13}{*}{subtopic} & \multirow{20}{*}{} & 27.91 & 72.79 & 39.62 & 39.66 & 99.17 & 54.26 & 63.16 & 23.98 & 33.39 & 42.42 \\
CEREC\textsubscript{exp} & & & 59.29 & 84.93 & 69.57 & 40.25 & 84.68 & 53.88 & 66.43 & 26.19 & 37.20 & 53.55 \\
 ECB+ & & & 68.72 & 91.75 & 78.38 & 47.96 & 90.10 & 62.10 & 71.66 & 30.45 & 42.33 & 60.94 \\
 ECB+METAm & & & 49.03 & 86.49 & 62.36 & 35.46 & 90.35 & 50.62 & 62.74 & 18.40 & 28.25 & 47.08 \\
FCC-T& & & 56.02 & 62.50 & 57.55 & 69.91 & 80.87 & 73.88 & 68.64 & 59.53 & 62.56 & 64.66 \\
GVC & & & 52.09 & 79.10 & 62.04 & 44.61 & 89.18 & 57.83 & 64.29 & 27.82 & 37.87 & 52.58 \\
HyperCoref\textsubscript{exp} & & & 9.59 & 27.35 & 13.25 & 56.86 & 94.81 & 69.90 & 71.92 & 43.83 & 53.66 & 45.61 \\
MEANTIME\textsubscript{eng}& & & 48.16 & 71.73 & 56.87 & 66.15 & 88.09 & 74.79 & 77.63 & 57.50 & 65.54 & 65.73 \\
NewsWCL50r & & & 70.01 & 92.75 & 79.75 & 35.48 & 89.33 & 50.35 & 70.40 & 13.87 & 23.09 & 51.06 \\
NIdent & & & 81.77 & 86.87 & 84.25 & 34.88 & 69.04 & 46.34 & 54.37 & 31.38 & 39.80 & 56.80 \\
NP4E\textsubscript{cd}& & & 82.91 & 85.46 & 84.16 & 41.03 & 70.31 & 51.82 & 51.01 & 39.57 & 44.57 & 60.18 \\
WEC-Eng & & & 83.57 & 94.12 & 87.66 & 80.09 & 99.65 & 86.45 & 90.02 & 69.42 & 74.95 & 83.02 \\
\hline
uCDCR & & & 57.42 & 77.99 & 64.62 & 49.36 & 87.13 & 61.02 & 67.69 & 36.83 & 45.27 & 56.97 \\
\hline
 ECB+ & \multirow{7}{*}{topic} & & 71.56 & 86.86 & 78.41 & 47.64 & 78.34 & 58.83 & 64.73 & 31.15 & 41.68 & 59.64 \\
 ECB+METAm & & & 51.89 & 80.26 & 62.99 & 34.74 & 81.79 & 48.50 & 58.26 & 18.13 & 27.49 & 46.33 \\
FCC-T& & & 61.45 & 65.92 & 63.61 & 36.05 & 44.19 & 39.71 & 26.14 & 15.71 & 19.63 & 40.98 \\
HyperCoref\textsubscript{exp} & & & 8.99 & 43.54 & 13.94 & 52.04 & 93.23 & 66.09 & 68.70 & 37.94 & 48.38 & 42.80 \\
MEANTIME\textsubscript{eng}& & & 70.70 & 73.94 & 72.28 & 49.84 & 59.19 & 54.12 & 50.76 & 42.88 & 46.49 & 57.63 \\
WEC-Eng & & & 83.25 & 83.83 & 83.38 & 69.39 & 49.44 & 53.52 & 40.76 & 40.50 & 38.73 & 58.54 \\
\hline
uCDCR & & & 57.97 & 72.39 & 62.43 & 48.28 & 67.70 & 53.46 & 51.56 & 31.05 & 37.07 & 50.99 \\
\hline
CD2CR & \multirow{19}{*}{subtopic} & \multirow{10}{*}{entity} & 27.91 & 72.79 & 39.62 & 39.66 & 99.17 & 54.26 & 63.16 & 23.98 & 33.39 & 42.42 \\
CEREC\textsubscript{exp} & & & 59.29 & 84.93 & 69.57 & 40.25 & 84.68 & 53.88 & 66.43 & 26.19 & 37.20 & 53.55 \\
 ECB+ & & & 68.08 & 92.21 & 78.09 & 45.77 & 91.31 & 60.53 & 71.58 & 28.73 & 40.71 & 59.77 \\
 ECB+METAm & & & 67.75 & 92.44 & 77.86 & 45.17 & 91.97 & 60.04 & 72.60 & 28.63 & 40.67 & 59.52 \\
HyperCoref\textsubscript{exp} & & & 14.25 & 29.19 & 17.75 & 64.63 & 97.61 & 76.27 & 76.71 & 51.73 & 60.60 & 51.54 \\
MEANTIME\textsubscript{eng}& & & 48.14 & 72.59 & 57.17 & 65.23 & 88.06 & 73.67 & 77.17 & 57.53 & 65.12 & 65.32 \\
NewsWCL50r & & & 77.01 & 94.11 & 84.66 & 41.70 & 90.99 & 56.57 & 75.00 & 21.44 & 33.13 & 58.12 \\
NIdent\textsubscript{en-cd} & & & 81.77 & 86.87 & 84.25 & 34.88 & 69.04 & 46.34 & 54.37 & 31.38 & 39.80 & 56.80 \\
NP4E\textsubscript{cd}& & & 82.91 & 85.46 & 84.16 & 41.03 & 70.31 & 51.82 & 51.01 & 39.57 & 44.57 & 60.18 \\
\cline{3-13}
uCDCR & & & 58.57 & 78.96 & 65.90 & 46.48 & 87.01 & 59.26 & 67.56 & 34.35 & 43.91 & 56.36 \\
\cline{3-13}
 ECB+ & & \multirow{9}{*}{event} & 68.04 & 93.93 & 77.95 & 49.76 & 91.88 & 62.54 & 71.02 & 32.19 & 42.81 & 61.10 \\
 ECB+METAm & & & 21.32 & 71.46 & 32.40 & 21.51 & 91.08 & 34.32 & 49.98 & 11.25 & 18.06 & 28.26 \\
FCC-T& & & 56.02 & 62.50 & 57.55 & 69.91 & 80.87 & 73.88 & 68.64 & 59.53 & 62.56 & 64.66 \\
GVC & & & 52.09 & 79.10 & 62.04 & 44.61 & 89.18 & 57.83 & 64.29 & 27.82 & 37.87 & 52.58 \\
HyperCoref\textsubscript{exp} & & & 10.07 & 27.85 & 13.89 & 52.70 & 93.87 & 66.26 & 68.25 & 38.93 & 48.84 & 43.00 \\
MEANTIME\textsubscript{eng}& & & 54.25 & 73.51 & 60.18 & 70.56 & 92.81 & 78.24 & 79.89 & 61.19 & 67.54 & 68.65 \\
NewsWCL50r & & & 47.36 & 91.32 & 61.83 & 14.60 & 91.65 & 23.95 & 38.62 & 2.37 & 4.46 & 30.08 \\
WEC-Eng & & & 83.57 & 94.12 & 87.66 & 80.09 & 99.65 & 86.45 & 90.02 & 69.42 & 74.95 & 83.02 \\
\hline
uCDCR & & & 49.09 & 74.22 & 56.69 & 50.47 & 91.37 & 60.43 & 66.34 & 37.84 & 44.63 & 53.92 \\
\hline
 ECB+ & \multirow{12}{*}{topic} & \multirow{5}{*}{entity} & 71.58 & 88.29 & 79.02 & 45.18 & 82.74 & 58.00 & 66.82 & 28.87 & 40.04 & 59.02 \\
 ECB+METAm & & & 71.56 & 88.82 & 79.23 & 44.49 & 83.59 & 57.62 & 67.23 & 28.37 & 39.54 & 58.79 \\
HyperCoref\textsubscript{exp} & & & 18.20 & 48.98 & 25.70 & 59.10 & 96.32 & 72.58 & 73.25 & 43.50 & 54.21 & 50.83 \\
MEANTIME\textsubscript{eng}& & & 74.46 & 81.77 & 77.95 & 43.87 & 62.49 & 51.55 & 53.79 & 32.87 & 40.81 & 56.77 \\
\cline{3-13}
uCDCR & & & 58.95 & 76.96 & 65.47 & 48.16 & 81.29 & 59.94 & 65.27 & 33.40 & 43.65 & 56.35 \\
\cline{3-13}
 ECB+ & & \multirow{7}{*}{event} & 69.98 & 86.01 & 76.87 & 50.50 & 75.64 & 59.59 & 63.89 & 34.33 & 43.64 & 60.03 \\
 ECB+METAm & & & 22.93 & 59.37 & 32.82 & 21.51 & 82.18 & 33.90 & 49.20 & 11.79 & 18.90 & 28.54 \\
FCC-T& & & 61.45 & 65.92 & 63.61 & 36.05 & 44.19 & 39.71 & 26.14 & 15.71 & 19.63 & 40.98 \\
HyperCoref\textsubscript{exp} & & & 4.45 & 25.06 & 6.77 & 49.37 & 93.24 & 63.52 & 67.02 & 36.21 & 46.26 & 38.85 \\
MEANTIME\textsubscript{eng}& & & 53.00 & 46.49 & 49.53 & 68.66 & 50.71 & 58.34 & 48.82 & 61.03 & 54.25 & 54.04 \\
WEC-Eng & & & 83.25 & 83.83 & 83.38 & 69.39 & 49.44 & 53.52 & 40.76 & 40.50 & 38.73 & 58.54 \\
\hline
uCDCR & & & 49.18 & 61.12 & 52.16 & 49.25 & 65.90 & 51.43 & 49.31 & 33.26 & 36.90 & 46.83 \\
\hline
\end{tabular}
\caption{Overview of baseline performance with all metrics.}
\label{tab:full_performance}
\end{table*}

\begin{table}[h]
\scriptsize
\centering
\begin{tabular}{{|p{0.3\linewidth}|p{0.55\linewidth}|}}
\hline
\textbf{Dataset} & \textbf{Topics} \\
\hline
CD2CR & 0 - Science \& Technology \\
\hline
CEREC\textsubscript{exp} & 0 - Emails \\
\hline
\multirow{43}{*}{\makecell{ECB+  \& \\ ECB+METAM \\ (in brackets)}} & 1 – Celebrity Rehab \\
 & (2) – Oscar Hosts \\
 & 3 – Prison Escapes \\
 & 4 – LA Deaths \\
 & (5) – Coaching Dismissals \\
 & 6 – Film Negotiations \\
 & 7 – Title Defenses \\
 & 8 – Bank Explosions \\
 & 9 – ESA Amendments \\
 & 10 – Contract Offers \\
 & 11 – Parliamentary Elections \\
 & (12) – Anti-Piracy Operations \\
 & 13 – Major Fires \\
 & 14 – Supermarket Fires \\
 & 16 – Assassinations \\
 & (18) – Office Shootings \\
 & 19 – Protest Riots \\
 & 20 – Qeshm Earthquakes \\
 & (21) – Hit-and-Runs \\
 & 22 – Workplace Murders \\
 & (23) – Climbing Fatalities \\
 & 24 – Jewelry Robberies \\
 & 25 – Injury Reserves \\
 & 26 – Mafia Deaths \\
 & 27 – Software Patches \\
 & 28 – Political Deaths \\
 & 29 – Playoff Qualification \\
 & 30 – Cable Disruptions \\
 & 31 – Scoring Records \\
 & 32 – Double Murders \\
 & 33 – Murder Trials \\
 & (34) – Federal Nominations \\
 & (35) – DUI Arrests \\
 & (36) – Polygamy Trials \\
 & (37) – Indonesian Earthquakes \\
 & (38) – Minor Earthquakes \\
 & (39) – Actor Replacements \\
 & (40) – Product Announcements \\
 & (41) – Camp Bombings \\
 & (42) – Smartphone Releases \\
 & (43) – Corporate Acquisitions \\
 & (44) – Corporate Acquisitions \\
 & (45) – Murder Convictions \\
\hline
FCC-T & 0 - Football \\
\hline
GVC & 0 - Gun violence \\
\hline
\multirow{35}{*}{\makecell{HyperCoref\textsubscript{exp}  \\ (topic ids = names)}} & ABC\_Univision \\
 & ABCNews \\
 & Author \\
 & Blotter \\
 & Business \\
 & Entertainment \\
 & GMA \\
 & Health \\
 & HealthCare \\
 & International \\
 & Lifestyle \\
 & MillionMomsChallenge \\
 & Misc \\
 & News \\
 & Nightline \\
 & OnCampus \\
 & Photos \\
 & Politic \\
 & Politics \\
 & PollingUnit \\
 & Primetime \\
 & Social\_Climber \\
 & Sport \\
 & Sports \\
 & Technology \\
 & Test \\
 & TheLaw \\
 & ThisWeek \\
 & Travel \\
 & Unit \\
 & US \\
 & WaterCooler \\
 & WhatWouldYouDo \\
 & WN \\
 & WNT \\ 
 \hline
\end{tabular}
\caption{Topic composition of uCDCR}
\label{tab:topics1}
\end{table}

\begin{table}[h]
\scriptsize
\centering
\begin{tabular}{|p{0.3\linewidth}|p{0.55\linewidth}|}
\hline
\textbf{Dataset} & \textbf{Topics} \\
 \hline
 \multirow{4}{*}{MEANTIME\textsubscript{eng}} & 0 - Airbus \\
 & 1 -   Apple \\
 & 2 - GM \\
 & 3 -   Stock \\
 \hline
 \multirow{10}{*}{NewsWCL50r} & 0 – Pompeo’s meeting in PRK \\
 & 1 – Comey   memos \\
 & 2 – PRK   suspends nuclear tests \\
 & 3 – DNC   sues Russia, Trump campaign \\
 & 4 – Macron   and Trump meeting \\
 & 5 – Planning   of Trump’s visit to UK \\
 & 6 – Migrant   caravan crosses into US \\
 & 7 – Delays   of US mail tariffs \\
 & 8 – Mueller’s   questions for Trump \\
 & 9 – Iran   nuclear files \\
 \hline
NIdent\textsubscript{en-cd}  \&  NP4E\textsubscript{cd} & 0 - Bomb, explosion, kidnap \\
 \hline
 \multirow{18}{*}{WEC-Eng} & 0 - Airliner Accident \\
 & 1 –   Awards \\
 & 2 –   Beauty Pageant \\
 & 3 –   Civilian Attack \\
 & 4 –   Concert \\
 & 5 –   Contest \\
 & 6 –   Earthquake \\
 & 7 –   Eruption \\
 & 8 –   Festival \\
 & 9 –   Flood \\
 & 10 –   Meetings \\
 & 11 –   News Event \\
 & 12 –   Oilspill \\
 & 13 –   Rail Accident \\
 & 14 –   Solar Eclipse \\
 & 15 –   Terrorist Attack \\
 & 16 –   Weapons Test \\
 & 17 –   Wildfire \\
 \hline
\end{tabular}
\caption{(cont'd) Topic composition of uCDCR}
\label{tab:topics2}
\end{table}

\end{document}